\documentclass[10pt,journal,compsoc]{IEEEtran}
\ifCLASSOPTIONcompsoc
\usepackage[nocompress]{cite}
\else
\usepackage{cite}
\fi
\ifCLASSINFOpdf
 \usepackage[pdftex]{graphicx}
\else
\usepackage[dvips]{graphicx}
\DeclareGraphicsExtensions{.eps}
\fi
\usepackage{amsmath}
\usepackage{amssymb}
\usepackage{amsthm}

\usepackage{array}
\usepackage[linesnumbered,ruled]{algorithm2e}

\theoremstyle{definition}

\let\oldnl\nl
\newcommand{\nonl}{\renewcommand{\nl}{\let\nl\oldnl}}

\renewcommand\thesection{\Alph{section}}

\usepackage{multirow}
\usepackage[font=scriptsize]{subfig}

\usepackage{epstopdf}
\usepackage{subfiles}
\usepackage{makecell}
\usepackage{soul}
\usepackage{color, xcolor}

\newcommand{\lyx}[1]{\textcolor{green}{#1}}

\usepackage{pifont}
\newcommand{\cmark}{\ding{51}\xspace}%
\newcommand{\xmark}{\ding{55}\xspace}%

\usepackage{verbatim}
\newcommand{\highlight}[1]{%
\colorbox{yellow}{$\displaystyle#1$}}
\soulregister\cite7 
\soulregister\eqref7
\usepackage{booktabs}
\begin{document}
	%
	\title{Towards Deviation-Robust Agent Navigation via  Perturbation-Aware Contrastive Learning}
	%
	%
	%
	%
	\author{Bingqian~Lin\IEEEauthorrefmark{1}, Yanxin Long\IEEEauthorrefmark{1}, Yi Zhu, Fengda Zhu, Xiaodan Liang\IEEEauthorrefmark{2}, Qixiang Ye,  Liang Lin
	\IEEEcompsocitemizethanks{
	\IEEEcompsocthanksitem 
	\IEEEauthorrefmark{1}Equal contribution\protect\\
	\IEEEcompsocthanksitem 
	\IEEEauthorrefmark{2}Corresponding author\protect\\	\IEEEcompsocthanksitem Bingqian Lin and Yanxin Long are with Shenzhen Campus of Sun Yat-sen University, Shenzhen, China. \protect\\
			E-mail: \{linbq6,longyx9\}@mail2.sysu.edu.cn.
		
 \IEEEcompsocthanksitem Xiaodan Liang is with Shenzhen Campus of Sun Yat-sen University, Shenzhen, China, and also with Dark Matter Inc. \protect\\
 E-mail: liangxd9@mail.sysu.edu.cn.
  
  \IEEEcompsocthanksitem  Liang Lin is with Sun Yat-sen University, Guangzhou, China.\protect\\
			E-mail: linliang@ieee.org.
		
			\IEEEcompsocthanksitem Yi Zhu, Qixiang Ye are with University of Chinese
	Academy of Sciences (UCAS), Beijing, China.
	\protect\\
	E-mail: zhu.yee@outlook.com;  qxye@ucas.ac.cn.
	\IEEEcompsocthanksitem Fengda Zhu is with Monash University, Melbourne, Australia.
	\protect\\
	E-mail: fengda.zhu@monash.edu.
	
	}
		}
	
	%
	%
	
	\markboth{IEEE TRANSACTIONS ON PATTERN ANALYSIS AND MACHINE INTELLIGENCE}%
	{Shell \MakeLowercase{\textit{et al.}}: Bare Demo of IEEEtran.cls for Computer Society Journals}
		\IEEEtitleabstractindextext{%
		\begin{abstract}

Vision-and-language navigation (VLN) asks an
agent to follow a given language instruction to navigate through a real 3D environment.
Despite significant advances, conventional VLN agents are  trained  typically  under disturbance-free environments and may easily fail in real-world navigation scenarios, since they are unaware of how to deal with various possible disturbances, such as sudden obstacles or human interruptions, which widely exist and may usually cause an unexpected route deviation. In this paper, we present a model-agnostic training paradigm, called  \textbf{Pro}gressive \textbf{Per}turbation-aware Contrastive Learning (PROPER) to enhance the generalization ability of existing VLN agents to the real world,
by requiring them to learn towards deviation-robust navigation. Specifically, a simple yet effective path perturbation scheme is introduced to implement the route deviation, with which the agent is required to still navigate successfully following the original instruction. 
Since directly enforcing the agent to learn perturbed trajectories may lead to insufficient and inefficient training, 
a progressively perturbed trajectory augmentation strategy is designed, where the agent can self-adaptively learn to navigate under perturbation with the improvement of its navigation performance for each specific trajectory. 
For encouraging the agent to well capture the difference brought by perturbation and adapt to both perturbation-free and perturbation-based environments, a perturbation-aware contrastive learning mechanism is further developed by contrasting perturbation-free trajectory encodings and perturbation-based counterparts. Extensive experiments on the standard Room-to-Room (R2R) benchmark  show that PROPER can benefit multiple state-of-the-art VLN baselines in perturbation-free scenarios. We further collect the perturbed path data to construct an introspection subset based on the R2R, called Path-Perturbed R2R (PP-R2R). The results on PP-R2R show unsatisfying robustness of popular VLN agents and the capability of PROPER in improving the navigation robustness under deviation.

		\end{abstract}
		
		\begin{IEEEkeywords}
			Vision-and-language navigation, navigation robustness, progressive training, contrastive learning
	\end{IEEEkeywords}}

	\maketitle

	\IEEEdisplaynontitleabstractindextext

	%
	\IEEEpeerreviewmaketitle
	


	%
		
\IEEEraisesectionheading{\section{Introduction}\label{sec:introduction}}

	%
	%
	%
	%


\IEEEPARstart{V}{ision-and-Language } Navigation (VLN) \cite{anderson2018vision} is a challenging task that requires an embodied agent to navigate through complex visual environment following  natural language instructions to reach the goal position. It has raised widely research interests in recent years since an instruction-following navigation agent is more flexible and practical in many real-world applications, such as personal assistants and in-home robots. Existing studies have made great progress in developing VLN agents by designing cross-modal alignment modules \cite{ma2019self,landi2019perceive}, data augmentation strategies \cite{fried2018speaker,tan2019learning} or efficient learning paradigms \cite{zhu2020vision,wang2019reinforced}. However, conventional VLN agents  are trained typically based on the assumption that their navigation environment is  disturbance-free. As a result, they often fail to generalize well to real-world navigation scenes due to the uncertainty of unseen visual observations and the existence of various possible disturbances, e.g., sudden obstacles or human interruptions.

\begin{figure}[t]
\begin{centering}
\includegraphics[width=0.95\linewidth]{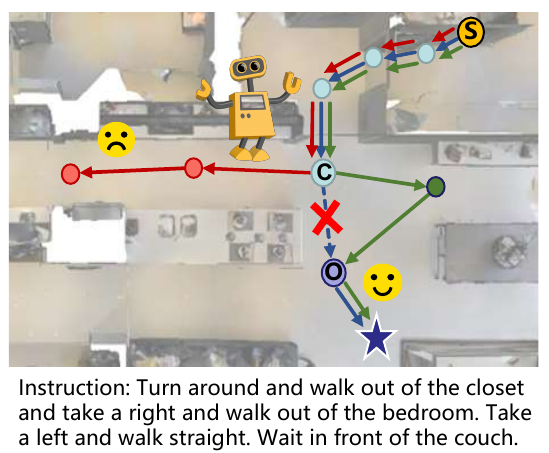}
\par\end{centering}
\vspace{-0.2cm}
\caption{
In real-world  scenarios, a VLN agent required to navigate from the start position $s$ to the goal position  may fail to move to $o$ from $c$ in the blue ground-truth (GT) trajectory due to a wrong action decision or possible disturbances and thus leads to a route deviation. The red and green  trajectories represent a failed and successful trajectory under deviation, respectively. 
}
\label{fig:motivation}
\vspace{-0.6cm}
\end{figure}

\begin{figure*}
	\centering
	\includegraphics[width=0.95\linewidth]{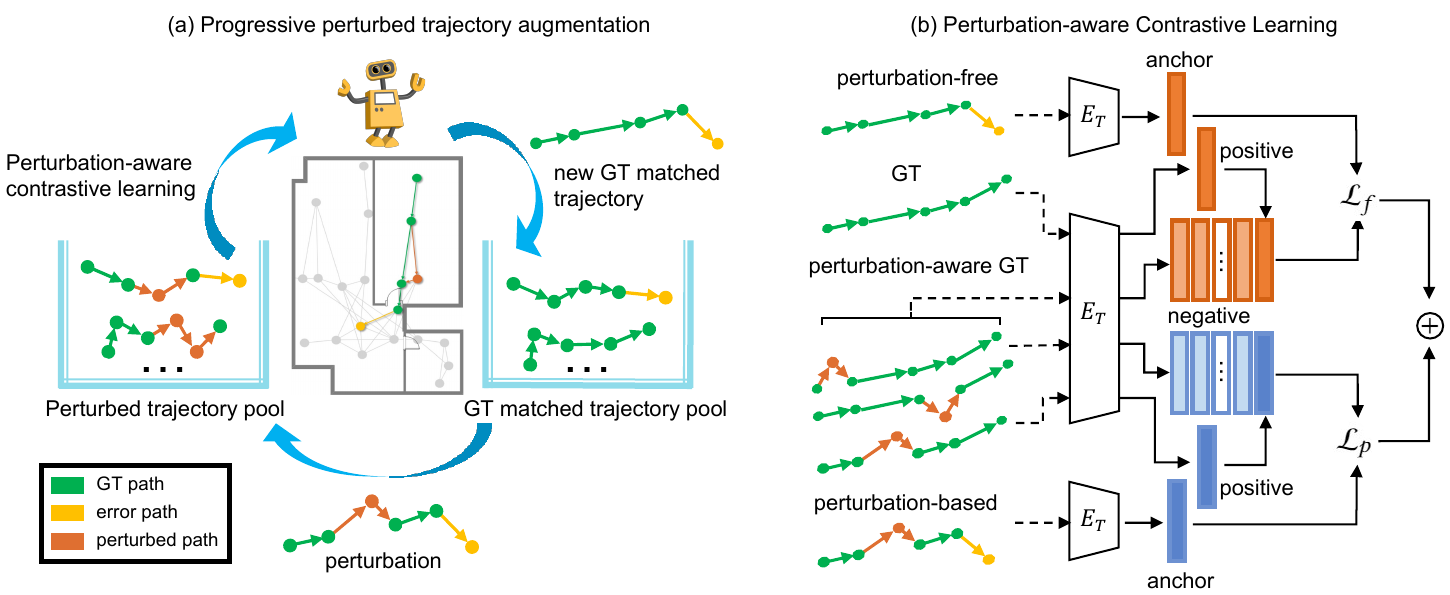}
	\caption{The overview of PROPER. (a) Progressively perturbed trajectory augmentation. At each training iteration, new GT matched trajectories are collected and imposed with perturbation. Then the new perturbed trajectories are combined with previous perturbed trajectories for training. (b) Perturbation-aware Contrastive Learning. 
	In perturbation-free and perturbation-based scenes, the anchor, positive and negative samples are obtained by the trajectory encoder $E_{T}$ for calculating the contrastive learning loss $\mathcal{L}_{f}$ and  $\mathcal{L}_{p}$, respectively.
	}
	
	\label{fig:comparison}
	\vspace{-0.4cm}
\end{figure*}


While real-world perturbations have different forms or appearances, for instance, the obstacles in the indoor environment may be a table or a chair, they usually cause an unexpected route deviation for a navigation trajectory. 
Moreover, a route deviation will also happen easily in disturbance-free scenes when a wrong action decision is made by the agent (Fig.~\ref{fig:motivation}).
Inspired by this, we consider: Can we improve the generalization ability of existing VLN agents towards real-world scenarios by simply requiring them to learn successful navigation under the unexpected route deviation?

In this paper, we make the first attempt to train deviation-robust VLN agents by introducing the unexpected perturbation during navigation, to make them more prepared for realistic navigation applications. Concretely, we propose a novel training paradigm, called \textbf{Pro}gressive \textbf{Per}turbation-aware Contrastive Learning (PROPER), where a simple yet effective path perturbation scheme via edge deletion is introduced to create the deviation, and the agent is encouraged to still navigate successfully under deviation following the original instruction. Such perturbation scheme can also be treated as an effective data augmentation technique, which does not need to rely on modeling the specific form or appearance of perturbation.

To enhance the deviation-robustness of the agent, a straightforward solution is to impose the perturbation into different trajectory instances directly. 
However, the learning ability and navigation performance of the agent for different trajectories are distinct and dynamically changed within the training.
Moreover, to enable robust action decision sequentially under both perturbation-based and perturbation-free environments, the agent also needs to well capture the difference brought by the perturbation.
To address these problems, a progressively perturbed trajectory augmentation strategy and a perturbation-aware contrastive learning mechanism are developed in PROPER.

With the agent's performance improvement during training, it will gradually learn to follow the ground-truth (GT) trajectories, leading to a progressive increase of GT matched trajectories. 
Based on this observation, we implement the progressively perturbed trajectory augmentation by firstly training the agent with perturbation-free trajectories. Then, as shown in Fig. \ref{fig:comparison}(a), we introduce the perturbed trajectories progressively by imposing the perturbation on the new GT matched trajectories in each training iteration, since they can  effectively indicate the adaption of the agent for the specific trajectory. 
By doing so, the agent can learn to navigate under perturbation in a self-adaptive way for high training efficiency.
The contrastive learning mechanism has been widely used in self-supervised learning areas recently \cite{chen2020improved,caron2020unsupervised,henaff2020data,bachman2019learning} and shown to be powerful for learning the relation among instance representations by contrasting positive and negative samples.
In our perturbation-aware contrastive learning paradigm, the agent learns distinct positive/negative relation among trajectories under different scenarios, as shown in Fig.~\ref{fig:comparison}(b).
Concretely, the agent treats the ground-truth (GT) trajectories as positive samples while perturbation-aware GT trajectories as negative samples when navigating without the perturbation. In perturbation-based scenes, however, the agent takes the perturbation-aware GT trajectory as positive sample rather than perturbation-free one. In this way, the agent can well capture the potential difference brought by perturbation and adapt to different scenarios.

We first conduct experiments on the standard VLN benchmark, i.e., the Room-to-Room (R2R) dataset \cite{anderson2018vision} to verify the ability of PROPER. 
Experimental results show that PROPER successfully improves perturbation-free navigation performance when plugged into multiple state-of-the-art VLN baselines, such as RelGraph \cite{hong2020language} and VLN$\circlearrowright$BERT \cite{Hong2021VLNBERTAR}, demonstrating that our introduced perturbed trajectories can be used as simple yet  powerful augmentation data combined with the proposed progressive training strategy and contrastive learning paradigm.
Furthermore, we collect the perturbed path data based on R2R to construct an introspection subset, dubbed as Path-Perturbed R2R (PP-R2R). We test the deviation-robustness of existing popular VLN agents \cite{tan2019learning, hong2020language,Hong2021VLNBERTAR} and the
agents trained with PROPER on PP-R2R. 
The results show poor generalization ability of these popular agents and the capability of PROPER in enhancing the  navigation robustness under deviation. 

To summarize, the main contributions of this paper are:
\begin{itemize}


\item{To the best of our knowledge, we take the first step to introduce the path perturbation during navigation for training deviation-robust VLN agents. These agents are therefore more appropriate for real-world applications where disturbances usually exist.} 


\item{A model agnostic training paradigm, i.e., Progressive Perturbation-aware Contrastive Learning (PROPER) is proposed, where  the path perturbation is implemented using edge deletion. A progressively perturbed trajectory augmentation strategy and a perturbation-aware contrastive learning mechanism are introduced in PROPER for efficient training under the perturbation\footnote{Note that PROPER is a training paradigm and therefore irrelevant to the validation/test setting, i.e., the agent trained using PROPER can be freely tested under both perturbation-free and perturbation-based environments.}.}

\item{Experimental results on the public R2R dataset and our constructed PP-R2R subset show the superiority of PROPER beyond state-of-the-art VLN baselines and its capability in enhancing navigation robustness under deviation. }

\end{itemize}

The remainder of this paper is organized as follows. Section \ref{related work} gives a brief review of the related work. Section \ref{method} describes the problem setup of VLN and then introduces our proposed method. Experimental results are provided in Section  \ref{experiment}. Section \ref{conclusion} concludes the paper and presents some outlook for future work.

\section{Related Work}
\label{related work}
\subsection{Vision-and-Language Navigation.}
Developing agents being capable of  understanding multi-modal information has received widely research interest in recent years~\cite{yin2020memory,yu2021learning,zhang2019more,das2019visual,gao2020hierarchical,wang2018fvqa}.
Vision-and-Language Navigation (VLN), proposed by Anderson et. al~\cite{anderson2018vision}, is a challenging multi-modal understanding task, which asks an agent to move across visual scenes to achieve the goal by following  given natural language instructions. 
Previous VLN works have explored various kinds of powerful paradigms and can be roughly divided into three categories:
1) Cross-modal alignment modules~\cite{ma2019self,landi2019perceive,qi2020object}: 
Ma \textit{et al.}~\cite{ma2019self} designed the visual-textual co-grounding module, which can use surrounding  observations to split the instructions into different phases and  identify current useful instruction information. Qi \textit{et al.}~\cite{qi2020object} classified important instruction information into object description and action specification to facilitate more efficient cross-modal alignment. 2) Efficient learning paradigms~\cite{wang2019reinforced,zhu2020vision,hao2020towards,li2019robust,majumdar2020improving,wang2020soft}: 
Zhu \textit{et al.}~\cite{zhu2020vision} designed multiple self-supervised auxiliary tasks to exploit rich semantic information in the environment. Hao \textit{et al.}~\cite{hao2020towards} and Majumdar \textit{et al.}~\cite{majumdar2020improving} largely enhanced the navigation performance by performing pretraining with the large-scale corpus following existing multi-modal pretraining approaches like~\cite{chen2020uniter,li2020unicoder,tan2019lxmert,lu2019vilbert}. 3) Data augmentation strategies~\cite{fried2018speaker,tan2019learning,fu2020counterfactual}: 
Tan \textit{et al.}~\cite{tan2019learning}
employed environmental dropout to generate augmented data by mimicking unseen environments. u \textit{et al.}~\cite{fu2020counterfactual}  proposed Adversarial Path Sampler for enhancing the quality of  augmented paths by using counterfactual thinking.



Although the above mentioned approaches have effectively improved the navigation performance, they all assume that the agents move in perturbation-free environments and seldom focus on the navigation robustness. In contrast, our work introduces the path perturbation during navigation and proposes a model-agnostic training paradigm for improving navigation robustness, which brings VLN agents closer to real-world scenarios. Moreover, we build a new dataset PP-R2R to serve as a helpful evaluation benchmark for verifying the robustness of VLN agents. 

\textbf{Data augmentation approaches in VLN.} Different from most existing data augmentation approaches in VLN that directly introduce the augmentation data for training~\cite{fried2018speaker,tan2019learning,fu2020counterfactual}, we adopt a progressive trajectory augmentation strategy in PROPER to enable the learning of augmentation trajectory in a self-adaptive way according to the agent's performance for each specific trajectory. As a result, the training efficiency can be effectively improved.

\subsection{Navigation Robustness.}

Robustness against possible perturbations is crucial for navigation agents deployed in real-world scenes. In vision-based robot navigation and autonomous driving areas, many works have been proposed for improving the navigation robustness. According to the type of the perturbation, here we simply divide these works into three categories: 1) obstacles~\cite{wang2020affordance,Moghadam2021AnAD,morad2021embodied,Cheng2022RealTimeTP,Kareer2022ViNLVN,Wenzel2021VisionBasedMR,Blum2022VisionbasedNA}: Wang \textit{et al.}~\cite{wang2020affordance} developed an affordance extraction procedure to enable the agent to navigate among movable obstacles. Wenzel \textit{et al.}~\cite{Wenzel2021VisionBasedMR} tackled the obstacle avoidance problem via a data-driven end-to-end deep learning approach. 2) collisions~\cite{giannakopoulos2021a,du2019group,niu2021accelerated,Wang2020SafeAE,Song2022SmoothTC,Li2020PredictionBasedRF,Shen2020CollisionAI}: Niu \textit{et al.}~\cite{niu2021accelerated} presented a sensor-level mapless collision avoidance algorithm for mobile robots to improve their collision avoidance ability in the real-world environment. 3) adversarial attacks~\cite{liu2020spatiotemporal,hamdi2020sada,Cao2022AdvDORA,Cheng2022PhysicalAO,Choi2022AdversarialAA,Zhang2022OnAR,Buddareddygari2021TargetedAO}: Cao \textit{et al.}~\cite{Cao2022AdvDORA} generated realistic adversarial trajectories and showed the strong attack ability of their method on autonomous vehicles. Although being widely studied in pure visual navigation, the navigation robustness has been largely ignored in VLN. Moreover, existing works in visual navigation tend to consider a specific kind of perturbation.

In this work, we aim to improve the robustness of VLN agents by asking them to explore another instruction-correlated trajectory for successful navigation after perturbation. This is more challenging generally than pure vision-based navigation where the agents only need to find a new trajectory to achieve the goal position. Through a simple edge deletion strategy to create the route deviation instead of relying on the specific type of perturbation, our method can be easily adapted to many possible disturbances which consistently cause a route deviation.

\subsection{Contrastive Learning.}
Contrastive learning is an emerging family of self-supervised learning approaches, which can effectively capture the relation among instance representations by minimizing/maximizing the distance between positive/negative sample pairs.
Many effective contrastive learning methods have been developed recently. In the image domain~\cite{chen2020improved,caron2020unsupervised,bachman2019learning,tian2019contrastive,tian2020what,chen2020a}, Chen \textit{et al.}~\cite{chen2020a} proposed a simple framework for contrastive learning of visual representations without requiring specialized architectures or a memory bank. Tian \textit{et al.}~\cite{tian2020what} devised unsupervised and semi-supervised frameworks to reduce the mutual information between views for learning powerful visual representations. In the natural language domain~\cite{Chi2021InfoXLMAI,wu2020unsupervised,fang2020cert,giorgi2020declutr,saunshi2019a,iter2020pretraining}, Fang \textit{et al.}~\cite{fang2020cert} presented CERT to conduct contrastive self-supervised learning at the sentence level for pretraining language models. In the graph domain~\cite{Zeng2021ContrastiveSL,hafidi2020graphcl,qiu2020gcc,hassani2020contrastive,You2020GraphCL,Suresh2021AdversarialGA}\, You \textit{et al.}~\cite{You2020GraphCL} designed different types of graph augments to conduct graph contrastive learning for learning unsupervised graph representations.

Most existing contrastive learning works are designed for static image, language or structural features, which are inappropriate for VLN where the useful instruction information and visual observations are dynamically changed during navigation. Moreover, most of them aim at the single scenario. In this work, we introduce a perturbation-aware contrastive learning mechanism, which learns the sequential trajectory encodings containing dynamic action information under different scenes. As a result, the agent can effectively capture the characteristics of perturbation-based and perturbation-free environments and adapt to both of them.

\section{Method}
\label{method}

In this section, we first review the problem setup of VLN in Sec. \ref{Original Problem Setup in VLN}. Then we detailedly describe our proposed PROPER, including perturbed trajectory construction in Sec. \ref{Perturbed Trajectory Construction}, progressively perturbed trajectory augmentation strategy in Sec. \ref{Perturbation-conditioned Navigation}, and perturbation-aware contrastive learning scheme in Sec. \ref{Perturbation-aware Contrastive Learning}.  

\subsection{Problem Setup}
\label{Original Problem Setup in VLN}
Given a language instruction $\mathbf{I}=\{w_{0},...,w_{L}\}$ with $L$ words, a VLN agent is required to find a route from a start viewpoint $c_{0}$ to the target viewpoint $c_{T}$.  At each timestep, $t$, the agent  observes a panoramic view, which contains 36 image views  $\{o_{t,i}\}_{i=1}^{36}$. Each image view $o_{t,i}$ includes an RGB image $b_{t,i}$ accompanied with its orientation ($\theta_{t,i}^{1}$,$\theta_{t,i}^{2}$), where $\theta_{t,i}^{1}$ and $\theta_{t,i}^{2}$ are the
angles of heading and elevation, respectively.  With the instructions and current visual observations, the agent infers the action for each step $t$ from the candidate actions list, which consists of $J_{t}$ neighbors of the current node in the navigation connectivity graph $\mathcal{G}=(V,E)$  and a stop action. $V$ and $E$ represent the nodes and edges in the navigation connectivity graph. Generally, the navigator is a sequence-to-sequence model with the encoder-decoder architecture \cite{anderson2018vision,thomason2019vision}. Following most existing VLN works \cite{tan2019learning,Hong2021VLNBERTAR, hong2020language}, we include both the imitation learning (IL) paradigm and reinforcement learning (RL) paradigm for training the agents.

Conventional VLN assumes that the agent can move freely across arbitrary predefined connected nodes. However, in real-world scenes, the navigation between two connected nodes may fail and thus cause a route deviation. Therefore, we propose to train  deviation-robust VLN agents by introducing the path perturbation during  navigation.

    
\begin{figure}[tb]
	\centering
	\includegraphics[width=0.95\linewidth]{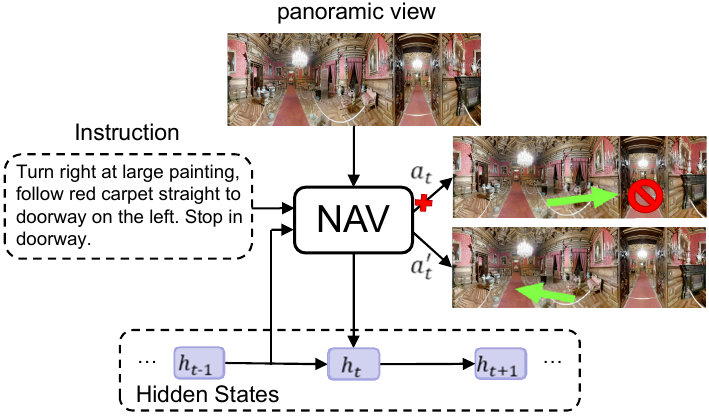}
	\vspace{-0.2cm}
	\caption{The flowchart of navigating under perturbation. At timestep $t$, the agent outputs the hidden state $h_{t}$ and the action $a_{t}$ based on the given instruction,  current panoramic view, and previous hidden state $h_{t-1}$. If a perturbation is conducted, the agent will make alternative action $a'_{t}$.  }
	
	\label{fig:flowchart}
	\vspace{-0.4cm}
\end{figure}

\subsection{Perturbed Trajectory Construction}
\label{Perturbed Trajectory Construction}
Since the real-world perturbations tend to have diverse forms or appearances while usually causing a route deviation, we  focus more on how to create a route deviation and train a robust VLN agent under the unexpected deviation directly in this paper. To this end, we introduce the perturbation based on the navigation connectivity graph by directly deleting the connectivity between two nodes to enforce a route deviation.
The flowchart is given in Fig.~\ref{fig:flowchart}.
Specifically, denote the current position and the next expected position during navigation as $c_{t}$ and $c_{t+1}$, respectively. The next action prediction of the agent is $a_{t}$. Then, the perturbation imposed on timestep $t$, denoted as $\mathrm{PE_{t}}$, is conducted by cutting the connectivity between $c_{t}$ and $c_{t+1}$:
\begin{align}
\mathrm{PE_{t}}= \mathrm{Del}(c_{t}, c_{t+1}),
\end{align}
where $\mathrm{Del}(a,b)$ means deleting the edge $ab$.  In principle, a valid edge deletion should not destroy the reachability to the goal position from the start point. As a result, the agent is enforced to make alternative action $a'_{t}$ to choose a new route to achieve the goal position.  

To facilitate the implementation of  perturbation-based navigation, we collect the deletable edges in advance. Moreover, if the perturbed edge matches the edge in the related ground-truth (GT) trajectory, the original GT trajectory will be inappropriate for supervision since the partial path in it is no longer available. In the following, we will describe in detail about the collection of deletable edges and the construction of perturbation-aware GT trajectories.

\textbf{Deletable Edge Collection.}
To ensure the perturbation to be general and diverse, we gather all ground-truth trajectories in R2R to collect the deletable edges. For each trajectory, the adjacent viewpoints are identified sequentially whether the direct edge between them can be removed while ensuring the connectivity between the original start position and target position. The edges satisfying such condition will be recorded for each ground-truth route in advance. The process of deletable edge collection is presented in Algorithm \ref{algorithm 1}. $OPA_{n}$ and $OPA$ represent the deletable edges related to specific ground-truth $p_{n}$ and whole dataset, respectively. $N^{p}$ is the total number of paths. $N_{n}^{v}$ denotes the total number of viewpoints in $p_{n}$. 

\begin{algorithm}[t]
	\SetAlgoLined
	\caption{Collection of deletable edges}
	\label{algorithm 1}
	{\small{
			\KwIn{ground-truth paths set $P=\{p_{n}\}_{n=1}^{N^{p}}$,  $p_{n}=\{v_{n,i}\}_{i=1}^{N_{n}^{v}}$, $N^{p}$, $N_{n}^{v}$.  }
			
	\KwOut{ $OPA$}	
			\BlankLine
		$OPA$ = \{\}\\
		\For{$p
		_{n}$ in $P$}{ 
		$OPA_{n}$ = \{\}\\
		\For{ adjacent point pairs $(v_{n,i}, v_{n,i+1})$ in $p_{n}$}{
		e=edge$(v_{n,i}, v_{n,i+1})$\\
		delete e\\
		\If {connectivity$(v_{n,i}, v_{n, N_{n}^{v}})$ is True}{

		$OPA_{n}$ $\leftarrow$ $OPA_{n}$ $\cup$ e
		}recover e\\
		}
			$OPA$ $\leftarrow$ $OPA$ $\cup$ $OPA_{n}$\\
			} 
		
	\Return  $OPA$
	}}

\end{algorithm} 

In R2R, the agent conducts the navigation action by choosing a node in the navigation graph. Due to the discrete characteristics of the R2R dataset, the agent may be enforced to move too far away from its current position when meeting the perturbation. As a result, it will be hard for the agent to find referred visual objects in the original instruction during deviation thus leading to redundant trajectories.
To mitigate this problem, an additional constraint is introduced, i.e., the distance between the alternatively chosen node after the edge deletion and the originally chosen node is smaller than a predefined threshold $r$. If the distances between all candidate alternative nodes and the originally chosen node are larger
than the threshold $r$, the deletion will not be
conducted. Considering that the visible scenes between neighbor nodes have a relatively large overlap and the distance between neighbor nodes is variable for different nodes and scenes, we set $r$ to be the average distance between neighbor nodes in a specific scene for reasonability. 

Through the distance constraint, the alternative chosen node can be seen as a neighbor of the originally chosen node. Therefore, the agent can also see the objects/scenes in the originally chosen node from the alternative chosen node. In this way, it can be effectively ensured that the instruction is relevant to the new perturbed path and the agent can navigate back according to the original instruction.

\begin{figure}[tb]
	\centering
	\includegraphics[width=0.95\linewidth]{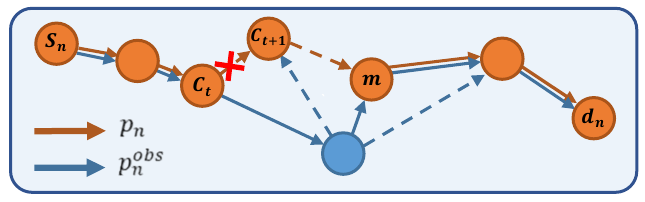}
	\vspace{-0.2cm}
	\caption{The illustration of the perturbation-aware GT trajectory construction. When  the edge $(c_{t},c_{t+1})$ in the GT trajectory $p_{n}$ is perturbed, the perturbation-aware GT trajectory $p_{n}^{obs}$ is constructed by connecting the sub-paths $(s_{n},c_{t})$, $(c_{t},m)$, and $(m,d_{n})$. $(s_{n},c_{t})$ and $(m,d_{n})$ overlap with the original $p_{n}$. $m$ is the end point of the shortest path whose beginning point is $c_{t}$ and the end point is on the sub-path $(c_{t+1},d_{n})$. }
	
	\label{fig:obs-gt}
	\vspace{-0.2cm}
\end{figure}

\textbf{Perturbation-aware GT Trajectory Construction.}
In R2R dataset, the ground-truth (GT) trajectory  is the shortest path following the given language instruction from the start to the target position. When the perturbation is just imposed on the edge in the GT trajectory, the original GT can not be treated as the optimal trajectory since the partial path snippets in it will not be available. Therefore, it is necessary to reconstruct perturbation-aware GT trajectories to serve as new supervision under perturbation.
Denote perturbation-free GT trajectory and perturbation-aware GT trajectory as $p_n$ and $p_n^{obs}$, respectively.
We construct $p_n^{obs}$ following two principles intuitively:  1) the additional exploration trajectory after the perturbation is as short as possible, 
2) Apart from the additional exploration trajectory due to the perturbation, other parts should be the same as those in $p_{n}$.
Fig. \ref{fig:obs-gt}  gives the illustration of constructing $p_n^{obs}$.  Specifically, denote the start viewpoint in $p_{n}$ as $s_{n}$, and the target position as $d_{n}$. 
The variables $c_{t}$ and $c_{t+1}$ are the beginning point and the end point of the deletable edge in the GT path $p_{n}$, respectively.
Then, the $p_{n}^{obs}$ is obtained by:
\begin{equation}
\begin{aligned}
p_{n}^{obs} = p_{n}(s_{n},c_{t})+p(c_{t}, m)
+p_{n}(m,d_{n}),
\end{aligned}
\end{equation}
where $p(a,b)$ denotes the sub-path beginning from $a$ to $b$, and $p_{n}(a,b)$ denotes the sub-path beginning from $a$ to $b$ in $p_{n}$. Denote  $\mathrm{L}(p(a,b))$ as the length of the sub-path $p(a,b)$. The variable $m$ is the end point of the shortest path whose beginning point is $c_{t}$ and the end point is on the sub-path $p_{n}(c_{t+1},d_{n})$: 
\begin{equation}
\begin{aligned}
    &\qquad \quad m=\mathrm{argmin}\;\mathrm{L}(p(c_{t},m)),\\
    s.t.\;&m\in p_{n}(c_{t+1},d_{n}),\;p(c_{t},m)\neq p_{n}(c_{t},c_{t+1}).
\end{aligned}
\end{equation}
The constraint $p(c_{t},m)\neq p_{n}(c_{t},c_{t+1})$ means excluding the perturbed edge $(c_{t},c_{t+1})$ since it is unavailable. Then $p(c_{t},m)$ and $ p_{n}(m,d_{n})$ can be used for encouraging the agent to navigate back as soon as possible after the perturbation and continue to follow the instruction to achieve the goal position, respectively.
By following the two principles to construct the perturbation-aware GT trajectory $p_{n}^{obs}$, we do not need to generate new instruction for $p_{n}^{obs}$ since the deviation is ensured to be as short as possible.  

\subsection{Progressively Perturbed Trajectory Augmentation}
\label{Perturbation-conditioned Navigation}
Due to different instruction/scene complexities and trajectory lengths, different  trajectories have distinct difficulties for the navigation agent.
Moreover, 
perturbation-based trajectory is harder essentially than its perturbation-free counterpart since the agent should navigate back to the correct instruction-related trajectory with it for ensuring successful navigation.
As a result, directly enforcing the agent to navigate under perturbation without considering its learning ability may cause insufficient and inefficient training.
In PROPER,
we propose a progressively perturbed trajectory augmentation strategy, where the agent's learning starts out with only perturbation-free trajectories and then adapts to perturbation-based trajectories.
This is implemented by imposing the perturbation on the new GT matched trajectories, i.e., the newly added trajectories which have partially or fully overlapped paths
with the GT trajectories in each training iteration, since the match to the GT trajectory can be viewed as a reliable indicator for the agent's adaption for the specific trajectory.
In this way, the agent can navigate under perturbation regarding its navigation capability for each specific trajectory to enhance its robustness more efficiently.

\begin{algorithm}[t]
	\SetAlgoLined
	\caption{Progressively Perturbed Trajectory Augmentation}
	\label{algorithm 2}
	{\small{
			\KwIn{Current iteration $n$, total number of iteration $N_{n}$, episode length $N_{t}$, perturbed data $\boldsymbol{M}_{n-1}^{p}$, instruction feature $\boldsymbol{I}$, visual feature $\boldsymbol{V}_{t}$, the navigator NAV with parameter $\pi_{n-1}$. }
			
	\KwOut{Optimized $\pi$}	
			\BlankLine
		\While{$n$ $<$ $N_{n}$}{
		$\boldsymbol{M}_{n}^{p}$ = \{\}\\
        \For{$t$= 0 : $N_{t}$}{
		\{$\boldsymbol{T}_{t}$\} $\leftarrow$ rollout($\{\boldsymbol{I}\}$, $\{\boldsymbol{V}_{t}\}$)\\
		$\boldsymbol{M}_{t}$ = match(\{$\boldsymbol{T}_{t}$\}, \{$\boldsymbol{T}_{\{GT\}}$\})\\
		$\boldsymbol{M}_{t}^{p}$ = Perturb($\boldsymbol{M}_{t}$)\\
		$\boldsymbol{M}_{n}^{p}$= $\boldsymbol{M}_{n}^{p}$ $\cup$ $\boldsymbol{M}_{t}^{p}$}

		$\boldsymbol{M}_{n-1}^{p}$ = $\boldsymbol{M}_{n-1}^{p}$ $\cup$ $\boldsymbol{M}_{n}^{p}$ \\
		$\pi_{n}$ $\leftarrow$ NAV($\boldsymbol{M}_{n-1}^{p}$, $\pi_{n-1}$) \\
		$\pi_{n-1}$ $\leftarrow$ $\pi_{n}$\\
}
		
	}}
\end{algorithm}

Algorithm \ref{algorithm 2} gives the procedure of the progressively perturbed trajectory augmentation process. Specifically, at each navigation timestep $t$ during the training iteration $n$, we get the navigation trajectories $\{\boldsymbol{T}_{t}\}$ in a minibatch under the instruction feature $\{\boldsymbol{I}\}$ and visual features $\{\boldsymbol{V}_{t}\}$. 
Then we select new GT matched trajectories $\boldsymbol{M}_{t}$ by $\mathrm{match}(\{\boldsymbol{T}_{t}\}, \{\boldsymbol{T}_{\{GT\}}\})$. 
The perturbation is implemented on   $\boldsymbol{M}_{t}$ to generate $\boldsymbol{M}_{t}^{p}$. When the navigation ends, we get the new perturbed trajectories  $\boldsymbol{M}_{n}^{p}$ for the current iteration $n$. Then these perturbed trajectories are added into the previous perturbed trajectories $\boldsymbol{M}_{n-1}^{p}$ and  introduced into training.

\subsection{Perturbation-aware Contrastive Learning}
\label{Perturbation-aware Contrastive Learning}
When training with both non-perturbed trajectories and progressively introduced perturbed trajectories, it is expected that the agent learns to adapt to the perturbation-based environment gradually and meanwhile improve (or keep) the navigation performance under perturbation-free scenarios. To this end, the agent should be able to capture the potential characteristics of these two different scenarios and the differences between them. For example, when navigating in the perturbation-free environment, the agent should learn to follow the original ground-truth (GT) trajectory. In perturbation-based scenarios, however, it should learn to follow perturbation-aware GT trajectory since the original GT trajectory is no longer available. Therefore, we propose a perturbation-aware contrastive learning paradigm during the progressive trajectory augmentation process, which enables the agent to distinguish positive/negative relations between trajectories regarding the two different scenarios for making robust action decision.

Specifically, we obtain the trajectory encoding at the end timestep of the navigation for contrastive learning. To simplify the implementation, we use the hidden state $h_{t}$ of the action decoder $E_{T}$ to serve as the trajectory encoding, following~\cite{wang2019reinforced, wang2020environment}. And the positive/negative samples are carefully selected for sufficient exploration of the relation between  trajectories to capture the characteristics of the two different scenarios. 

\begin{table}[t]
	\centering
	\fontsize{7}{7}\selectfont
	\caption{Notations of trajectory encodings in the perturbation-aware contrastive learning paradigm. 
	}\label{tab:notations}
		\resizebox{1.0\linewidth}{!}{
	{\renewcommand{\arraystretch}{1.2}	\begin{tabular}{cc}
				\specialrule{.1em}{.05em}{.05em}	
			
	Trajectory encoding&Notations\\ \hline
	perturbation-free&$\mathbf{e}_{f}$\\ 

	perturbation-based&$\mathbf{e}_{p}$\\
		GT&$\mathbf{e}_{g}$\\
	perturbation-aware GT (same perturbation position with $\mathbf{e}_{p}$)&$\mathbf{e}_{og}$ \\
    perturbation-aware GT (different perturbation position from $\mathbf{e}_{p}$)&$\mathbf{e}_{og'}$  \\
	others in the same minibatch &$\mathbf{e}_{m}$\\
	\specialrule{.1em}{.05em}{.05em}	\end{tabular}}}
\vspace{-0.2cm}	
\end{table}

\begin{figure}[tb]
	\centering
	\includegraphics[width=0.95\linewidth]{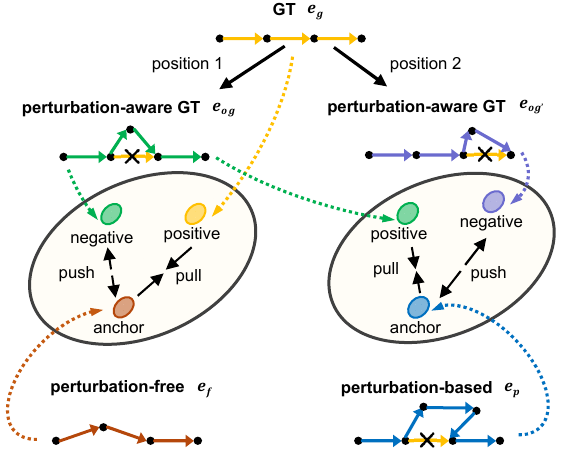}
	\caption{The illustration of perturbation-aware contrastive learning under different scenarios. 
	For perturbation-free trajectory encoding $\boldsymbol{e}_{f}$, the positive sample and the intra-negative sample are GT trajectory encoding $\boldsymbol{e}_{g}$ and perturbation-aware GT trajectory encoding $\boldsymbol{e}_{og}$, respectively. For perturbation-based trajectory encoding $\boldsymbol{e}_{p}$, the positive sample and the intra-negative sample are perturbation-aware GT trajectory encoding $\boldsymbol{e}_{og}$ and the perturbation-aware GT trajectory encoding of different perturbation position   $\boldsymbol{e}_{og'}$, respectively. For simplicity, we omit the inter-negative samples $\boldsymbol{e}_{m}$ in both two scenarios in the figure,  which are other trajectories in the same minibatch.  
	}
	\label{fig:contrastive}
	\vspace{-0.2cm}
\end{figure}

\textbf{Positive/Negative Sample Selection.}
The perturbation-aware contrastive learning paradigm is conducted under both perturbation-free and perturbation-based scenarios. We first give the notations of trajectory encodings in Table~\ref{tab:notations}. And the illustration of the positive/negative sample selection under different scenes is given in Fig. \ref{fig:contrastive}.  
Denote the perturbation-free trajectory encoding and perturbation-based trajectory encoding as $\boldsymbol{e}_{f}$ and $\boldsymbol{e}_{p}$, respectively. For $\boldsymbol{e}_{f}$, the GT trajectory encoding $\boldsymbol{e}_{g}$ is selected as the positive sample. While for $\boldsymbol{e}_{p}$, the perturbation-aware GT trajectory encoding $\boldsymbol{e}_{og}$ rather than $\boldsymbol{e}_{g}$ is chosen as the positive sample. In this way, the agent can learn to imitate the optimal trajectory under different scenarios for better action decision. 

We introduce two kinds of negative samples, i.e., intra-negative samples and inter-negative samples, to conduct contrastive learning. For a specific trajectory $\boldsymbol{e}$, we treat the trajectory that only has partial non-overlap path snippet with $\boldsymbol{e}$ as the intra-negative sample. And the trajectory which is totally different from $\boldsymbol{e}$ is viewed as the inter-negative sample. For the perturbation-free trajectory encoding $\boldsymbol{e}_{f}$, we choose all perturbation-aware GT trajectories $\boldsymbol{e}_{og}$ as intra-negative samples. While for perturbation-based trajectory encoding $\boldsymbol{e}_{p}$, 
the perturbation-aware GT trajectories of other perturbation positions for the same trajectory, denoted as $\boldsymbol{e}_{og'}$, are selected as intra-negative samples. The intra-negative samples in both two scenarios serve as hard negative samples for better distinguishing  the positive/negative sample since they only have small difference from the positive samples, as shown in Fig.~\ref{fig:contrastive}. 

Following many existing contrastive learning works~\cite{chen2020a,kim2020adversarial}, we use different instances in a minibatch, which are denoted as $\boldsymbol{e}_{m}$, as inter-negative samples under both perturbation-free and perturbation-based environments  to further encourage the sufficient relation exploration among different trajectories.

\textbf{Contrastive Losses for Different Scenes.}
We use the InfoNCE loss in the proposed contrastive learning paradigm. Specifically,
The contrastive loss $\mathcal{L}_{f}$ in perturbation-free environment is calculated by:
\begin{equation}
\begin{aligned}
\label{eq:contrastive loss}
\mathcal{L}_{f}=&-\mathrm{log}\frac{\mathrm{exp}(\mathrm{sim}(\boldsymbol{e}_{f}, \boldsymbol{e}_{g})/\tau)}{\mathrm{exp}(\mathrm{sim}(\boldsymbol{e}_{f}, \boldsymbol{e}_{g})/\tau)+\sum\limits_{\boldsymbol{e}_{og}}\mathrm{exp}(\mathrm{sim}(\boldsymbol{e}_{f}, \boldsymbol{e}_{og})/\tau)} \\
&-\mathrm{log}\frac{\mathrm{exp}(\mathrm{sim}(\boldsymbol{e}_{f}, \boldsymbol{e}_{g})/\tau)}{\mathrm{exp}(\mathrm{sim}(\boldsymbol{e}_{f}, \boldsymbol{e}_{g})/\tau)+\sum\limits_{\boldsymbol{e}_{m}}\mathrm{exp}(\mathrm{sim}(\boldsymbol{e}_{f}, \boldsymbol{e}_{m})/\tau)}, 
\end{aligned}
\end{equation}
where $\tau$ is the temperature parameter \cite{chen2020a}. $\mathrm{sim}(a, b)$ represents the cosine similarity between two features $a$ and $b$. In perturbation-based scene, the contrastive loss $\mathcal{L}_{p}$ is obtained by:
\begin{equation}
\begin{aligned}
\mathcal{L}_{p}=&-\mathrm{log}\frac{\mathrm{exp}(\mathrm{sim}(\boldsymbol{e}_{p}, \boldsymbol{e}_{og})/\tau)}{\mathrm{exp}(\mathrm{sim}(\boldsymbol{e}_{p}, \boldsymbol{e}_{og})/\tau)+\sum\limits_{\boldsymbol{e}_{og'}}\mathrm{exp}(\mathrm{sim}(\boldsymbol{e}_{p}, \boldsymbol{e}_{og'})/\tau)} \\
&-\mathrm{log}\frac{\mathrm{exp}(\mathrm{sim}(\boldsymbol{e}_{p}, \boldsymbol{e}_{og})/\tau)}{\mathrm{exp}(\mathrm{sim}(\boldsymbol{e}_{p}, \boldsymbol{e}_{og})/\tau)+\sum\limits_{\boldsymbol{e}_{m}}\mathrm{exp}(\mathrm{sim}(\boldsymbol{e}_{p}, \boldsymbol{e}_{m})/\tau)}, 
\end{aligned}
\end{equation}

\textbf{Training Objectives.}
We train the agents using imitation learning (IL), reinforcement learning (RL), and the proposed contrastive learning paradigms.
In IL, the agent outputs the action prediction probability $p_{t}$, and takes the teacher action $a_{t}^{*}$ at each step $t$ to efficiently  imitate the expert trajectory. The loss of IL is:
\begin{align}
\mathcal{L}_{IL} =  \sum_{t}-a_{t}^{*}log(p_{t}).
\end{align}
The RL loss is calculated by:
\begin{align}
\mathcal{L}_{RL} =  \sum_{t}-a_{t}log(p_{t})A_{t},
\end{align}
where $A_{t}$ is the advantage in A2C algorithm \cite{tan2019learning}. 
Within the contrastive learning in the training process, the training objective of our PROPER is:
\begin{align}
\label{loss}
\mathcal{L} = \mathcal{L}_{RL}+\lambda_{1}\mathcal{L}_{IL}+\lambda_{2}\mathcal{L}_{f}+\lambda_{3}\mathcal{L}_{p},
\end{align}
where $\lambda_{1}$, $\lambda_{2}$, and $\lambda_{3}$ are the loss weights to balance the loss
items.

\section{Experiments}
\label{experiment}
In this section, we first introduce the experimental setup, such as datasets, evaluation metrics and implementation details in Sec. \ref{Experimental Setup}. The quantitative results on R2R and PP-R2R are given in Sec. \ref{Quantitative Results-VLN} and Sec. \ref{Quantitative Results-OVLN}, respectively. And the visualization results are presented in Sec. \ref{Qualitative Results}. 
\subsection{Experimental Setup}
\label{Experimental Setup}
\subsubsection{Datasets}
We conduct experiments on two datasets, i.e., the public R2R dataset \cite{anderson2018vision} and our constructed introspection subset, Path-perturbed R2R (PP-R2R) to validate our proposed method. Specifically, R2R~\cite{anderson2018vision} includes 10,800 panoramic views and 7,189 trajectories. Each panoramic view has 36 images and each trajectory is paired with three natural language instructions. 

\begin{table}[tb]
	\centering
		\fontsize{20}{20}\selectfont
		\caption{The  data statistics for the training and validation split of R2R and PP-R2R. Tra., Ste., and Dis. represent trajectory, step number and distance, respectively. For distance and step number, average value is reported.
		}\label{tab:datasets}
	\resizebox{1.0\linewidth}{!}{
	{\renewcommand{\arraystretch}{1.2}

		\begin{tabular}{c||c|c|c|c|c|c|c|c|c}
		\specialrule{.1em}{.05em}{.05em}
			\multirow{2}{*}{Datasets}&\multicolumn{3}{c|}{Train}&\multicolumn{3}{c|}{Val Seen}&\multicolumn{3}{c}{Val UnSeen}\cr\cline{2-10}
			&Tra.&Ste.&Dis.&Tra.&Ste.&Dis.&Tra.&Ste.&Dis.
			\\\hline			R2R&4,675&6.0&9.9&340&6.1&10.2&783&6.0&9.5\\
		PP-R2R&4,623&7.8&13.3&335&7.9&14.6&769&9.0&15.9\\
			\specialrule{.1em}{.05em}{.05em}
		\end{tabular}}}
\vspace{-0.2cm}	
\end{table}

\textbf{Our PP-R2R.} To construct PP-R2R, we collect the trajectory instances which can be imposed with perturbation in both training and validation sets of the R2R dataset. The comparison of data statistics is given in Table~\ref{tab:datasets}. From Table~\ref{tab:datasets} we can find that most trajectories in training and validation set of original R2R dataset can be imposed with perturbation, showing the feasibility of our proposed perturbation method.
Since the perturbation-aware GT trajectory has the same start point and end point as the original GT while partial path snippet different from the original GT, the perturbation-aware GT trajectory in PP-R2R is definitely longer than the paired original GT in R2R which is the shortest path between the predefined start point and end point~\cite{anderson2018vision}.
One should note that since the construction principle of perturbation-aware GT trajectory is to make a short deviation and follow the original instruction after perturbation, the difference between perturbation-aware GT paths and original ones  would not be too big naturally.
Moreover, while the statistics difference between them in Table~\ref{tab:datasets} seems small, the perturbation would cause significant effects as shown in Table~\ref{tab:results on obstacle-VLN}, i.e., a significant performance drop ($\sim$20 points for SR and SPL) when testing perturbation-free trained agents under perturbation-based environments compared to testing in perturbation-free environments.

We further collect some data statistics of the constructed PP-R2R to verify the generalization ability of our introduced perturbation. Specifically, we first calculate the minimum, maximum and average numbers of deletable edges in trajectories, which are 1, 6, and 4, respectively, showing that multiple perturbation-available positions exist in the trajectories generally.
Then, based on the calculated minimum and maximum numbers of deletable edges, we investigate the trajectory proportions in different number phases, which are shown in Table~\ref{tab:datasets-1}. From Table~\ref{tab:datasets-1} we can find that most trajectories have larger than 2 perturbation-available edges, which shows that diverse perturbed trajectories can be generated in PP-R2R. 

\begin{table}[t]
	\centering
	\fontsize{7}{7}\selectfont
	\caption{The proportions of trajectories which have different numbers of deletable edges. }\label{tab:datasets-1}
		\resizebox{0.9\linewidth}{!}{
	{\renewcommand{\arraystretch}{1.2}	\begin{tabular}{c||c|c|c}
				\specialrule{.1em}{.05em}{.05em}	
			Splits&1-2&2-4&4-6
	\\\hline			train&12.72\%&51.31\%&35.97\%\\
	Val Seen&11.94\%&48.06\%&40.00\%\\
	Val UnSeen&14.43\%&54.10\%&31.47\%\\
	\specialrule{.1em}{.05em}{.05em}	\end{tabular}}}
\end{table}

\begin{table}[t]
	\centering
	\fontsize{7}{7}\selectfont
	\caption{The proportions of trajectories whose perturbation can be imposed in different parts. ``Beginning'', ``Middle'', ``End'' denote the beginning part, the middle part, and the end part of a trajectory.  }\label{tab:datasets-2}
	\resizebox{0.9\linewidth}{!}{
	{\renewcommand{\arraystretch}{1.2}
		\begin{tabular}{c||c|c|c}
			\specialrule{.1em}{.05em}{.05em}	Splits&Beginning&Middle&End
	\\\hline			train&85.59\%&83.13\%&91.97\%\\
	Val Seen&84.73\%&81.44\%&94.90\%\\
	Val UnSeen&84.90\%&81.25\%&89.46\%\\
	\specialrule{.1em}{.05em}{.05em}	\end{tabular}}}
\vspace{-0.2cm}	
\end{table}

Intuitively, the perturbation imposed on different parts in trajectories also brings different impacts for navigation. Let ``Beginning'', ``Middle'', and ``End'' denote the beginning part, the middle part, and the end part of a trajectory, respectively.\footnote{For example, if a trajectory contains 6 nodes, then the ``Beginning'', ``Middle'', and ``End'' contain  nodes 1-2,  nodes 3-4, and  nodes 5-6, respectively.} We investigate how many trajectories whose perturbation can be introduced in ``Beginning'', ``Middle'', and ``End'', respectively, to verify the flexibility of the introduced perturbation. The statistics are presented in Table~\ref{tab:datasets-2}. From Table~\ref{tab:datasets-2} we can observe that most trajectories in both training and validation sets can be imposed with perturbation in different positions in the trajectory, showing the flexibility of our introduced perturbation.


\subsubsection{Evaluation Metrics}
Following most existing VLN works \cite{anderson2018vision,tan2019learning,Hong2021VLNBERTAR,hong2020language}, we use four popular metrics \cite{anderson2018vision} for the navigation performance evaluation on both R2R and our PP-R2R datasets: 1) Trajectory Length (TL) calculates the average length of the trajectory, 2) Navigation Error (NE) is the distance between target viewpoint  and agent stopping position. 3) Success Rate (SR) calculates the success rate of reaching the goal. 4) Success rate weighted by Path Length (SPL) makes the trade-off between SR and the trajectory length.

\subsubsection{Baselines}
Since our proposed PROPER is model-agnostic, we choose three popular VLN approaches, i.e., EnvDropout~\cite{tan2019learning}, RelGraph~\cite{hong2020language} and VLN$\circlearrowright$BERT~\cite{Hong2021VLNBERTAR}, as baselines to verify the effectiveness of it on both R2R and PP-R2R. These three baselines have representative model architectures: 1) EnvDropout is a sequence-to-sequence-based model, 2) RelGraph is a graph-based model, and 3) VLN$\circlearrowright$BERT is a Transformer-based model. Moreover, RelGraph and VLN$\circlearrowright$BERT are two recent state-of-the-art VLN approaches.

\subsubsection{Implementation Details}
For  selecting the new action $a'_{t}$ in perturbed trajectories,  we mask the original predicted action and sample a new one based on the action prediction probability. This potentially encourages the prediction of good actions with sub-maximum prediction probability, and thus benefits navigation since there exists uncertainty during RL action sampling. 
Following most existing VLN works~\cite{tan2019learning,hong2020language,Hong2021VLNBERTAR}, $\lambda_{1}$ for $\mathcal{L}_{IL}$ is set to 0.2. $\lambda_{2}$ and $\lambda_{3}$ for $\mathcal{L}_{f}$ and $\mathcal{L}_{p}$ are set to 1  empirically.
For training PROPER on R2R, when the baselines are EnvDropout~\cite{tan2019learning} and RelGraph~\cite{hong2020language}, we adopt the same two-stage procedure as them, i.e., first pretraining and then finetuning. The iterations for pretraining (80k for both EnvDropout and RelGraph) and finetuning (200k for EnvDropout and 300k for RelGraph) are kept the same as the baseline models. 
When the baseline is  VLN$\circlearrowright$BERT~\cite{Hong2021VLNBERTAR}, we adopt the same training setting and iterations as the baseline, i.e., directly finetuning the agent from released pretrained weights\footnote{https://github.com/YicongHong/Recurrent-VLN-BERT} for 300k iterations. When testing on PP-R2R, each baseline and its PROPER counterpart are trained for 80k iterations for fair comparison. The training  optimizers are kept the same as the baseline models.  
We use both the original training data and augmentation instruction-path pairs for training in R2R as the baseline models.

\begin{table*}[!htb]
	\fontsize{9}{9}\selectfont
	\caption{Comparison of single-run performance with the state-of-the-art methods on R2R. For baseline methods trained with PROPER (e.g., EnvDropout+PROPER), the agents are trained under the perturbation-based setting while tested under the perturbation-free environments. We report the metrics of TL (m), NE (m), SR (\%) and SPL (\%). Following \cite{Hong2021VLNBERTAR,hong2020language}, we also report the results with two digits for SR and SPL.}
	\label{tab:com with sota}
		\resizebox{1.0\linewidth}{!}{
	{\renewcommand{\arraystretch}{1.2}
		\begin{tabular}{c||c|c|c|c|c|c|c|c|c|c|c|c}
			\specialrule{.1em}{.05em}{.05em}
			\multirow{2}{*}{Method}&\multicolumn{4}{c|}{Val Seen }&\multicolumn{4}{c|}{Val Unseen}&\multicolumn{4}{c}{Test Unseen}\cr\cline{2-13}
			&TL&NE $\downarrow$&SR $\uparrow$&SPL $\uparrow$&TL&NE $\downarrow$&SR $\uparrow$&SPL $\uparrow$&TL&NE $\downarrow$&SR $\uparrow$&SPL $\uparrow$\cr
			\hline
			
        
            Seq2Seq \cite{anderson2018vision}&11.33&6.01&39&-&8.39&7.81&22&-&8.13&7.85&20&18\\
            Speaker-Follower \cite{fried2018speaker}&-&3.36&66&-&-&6.62&35&-&14.82&6.62&35&28\\
            RCM+SIL(train) \cite{wang2019reinforced}&10.65&3.53&67&-&11.46&6.09&43&-&11.97&6.12&43&38\\
            
            Regretful \cite{ma2019the}&-&3.23&69&63&-&5.32&50&41&13.69&5.69&48&40\\
       
        PREVALENT~\cite{hao2020towards} & 10.32 & 3.67 & 69 & 65 & 10.19 & 4.71 & 58 & 53 & 10.51 & 5.30 & 54 & 51 \\
            AuxRN \cite{zhu2020vision}&-&3.33&70&67&-&5.28&55&50&-&5.15&55&51\\
           \hline 
           EnvDropout \cite{tan2019learning}&11.00&3.99&62&59&10.70&5.22&52&48&11.66&5.23&51&47\\ EnvDropout+PROPER&11.67&\textbf{3.34}&\textbf{69}&\textbf{66}&14.68&\textbf{4.93}&\textbf{55}&\textbf{50}&14.74&\textbf{4.90}&\textbf{55}&\textbf{51}\\
             \hline
            RelGraph \cite{hong2020language}&10.13&3.47&67&65&9.99&4.73&57&53&10.29&4.75&55&52\\ RelGraph+PROPER&10.31&\textbf{3.21}&\textbf{70}&\textbf{67}&11.52&\textbf{4.68}&\textbf{58}&\textbf{53}&11.67&\textbf{4.56}&\textbf{58}&\textbf{53}\\
           \hline
            
		 VLN$\circlearrowright$BERT \cite{Hong2021VLNBERTAR}&11.13&2.90&72&68&12.01&3.93&63&57&12.35&\textbf{4.09}&\textbf{63}&57\\
            VLN$\circlearrowright$BERT+PROPER&11.07&\textbf{2.69}&\textbf{75}&\textbf{71}&12.09&\textbf{3.91}&\textbf{64}&\textbf{58}&12.67&4.11&\textbf{63}&\textbf{58}\\
		 \specialrule{.1em}{.05em}{.05em}

			\end{tabular}}}
	\vspace{-0.2cm}
\end{table*}

\begin{table}[!htb]
	\fontsize{12}{12}\selectfont
	\caption{Ablation study of PROPER in R2R. ``Per-free'' and ``Per-based'' represent perturbation-free and perturbation-based environments, respectively. The agents are evaluated on the validation set under the perturbation-free environments. The baseline is EnvDropout~\cite{tan2019learning}.}
	\label{tab:effectiveness of contrastive}

			\resizebox{1.0\linewidth}{!}{
	{\renewcommand{\arraystretch}{1.2}\begin{tabular}{c||c|c|c|c|c|c|c|c}
			 \specialrule{.1em}{.05em}{.05em}
			
			\hline
			\multirow{2}{*}{Method}&\multicolumn{4}{c|}{Training}&\multicolumn{4}{c}{Validation (Per-free)}\cr\cline{2-9}
			&\multicolumn{2}{c|}{Per-free}&\multicolumn{2}{c|}{ Per-based}&\multicolumn{2}{c|}{Val Seen}&\multicolumn{2}{c}{Val Unseen}\cr\cline{2-9}
			&ML&CL&ML&CL&NE &SR &NE &SR\cr
			\hline
			
			
          Teacher&-&-&-&-&4.68&56.2&6.10&42.9\\
          Teacher2Student&-&-&-&-&3.99&59.9&5.55&45.6\\
         \hline Teacher+RL&\cmark&&&&4.08&60.6&5.50&47.6\\
       Non-pro Aug&\cmark&&\cmark&&4.07&61.1&5.38&49.2\\ Progressive Aug&\cmark&&\cmark&&3.97&62.6&5.26&50.4\\ 
         
       
          PROPER&\cmark&\cmark&\cmark&\cmark&\textbf{3.90}&\textbf{63.3}&\textbf{5.13}&\textbf{51.3}\\
			
      	 \specialrule{.1em}{.05em}{.05em}
		\end{tabular}}}
		\vspace{-0.2cm}
\end{table}

\begin{table*}[t]
	\fontsize{8}{8}
	\selectfont
	\caption{The navigation performance of popular VLN agents on PP-R2R. All methods are trained on R2R training set for evaluating the generalization ability. ``Per-free'' and ``Per-based'' represent perturbation-free and perturbation-based environments, respectively. Bold fonts indicate our method.}
	\vspace{-0.2cm}
	\label{tab:results on obstacle-VLN}
		\resizebox{1.0\linewidth}{!}{
	{\renewcommand{\arraystretch}{1.2}\begin{tabular}{c||c|c|c|c|c|c|c|c}
		\specialrule{.1em}{.05em}{.05em}
			\multirow{2}{*}{Method}&\multicolumn{2}{c|}{Setting}&\multicolumn{3}{c|}{Val Seen }&\multicolumn{3}{c}{Val Unseen}\cr\cline{2-9}
			&PROPER&Validation&NE(m) $\downarrow$&SR(\%) $\uparrow$&SPL(\%)  $\uparrow$&NE(m) $\downarrow$&SR(\%) $\uparrow$&SPL(\%)  $\uparrow$\cr
			\hline
            \multirow{3}{*}{EnvDropout \cite{tan2019learning}}&\multirow{2}{*}{\xmark}&Per-free&4.65&54.3&51.5&5.84&45.9&42.4 \\
       
     
     &&Per-based&7.03&36.5&34.9&8.59&23.9&22.2\\ \cline{2-9}
      &\cmark&Per-based&\textbf{6.26}&\textbf{43.6}&\textbf{40.4}&\textbf{7.91}&\textbf{30.3}&\textbf{25.7}\\ 
        \hline       	
	\multirow{3}{*}{RelGraph \cite{hong2020language}}&\multirow{2}{*}{\xmark}&Per-free&4.23&60.6&58.2&5.05&52.5&49.3\\
 &&Per-based&7.02&37.7&36.4&8.17&27.5&25.7\\ \cline{2-9}
 &\cmark&Per-based&\textbf{6.42}&\textbf{40.7}&\textbf{36.8}&\textbf{7.47}&\textbf{31.7}&\textbf{27.3}\\ \hline
     \multirow{3}{*}{VLN$\circlearrowright$BERT \cite{Hong2021VLNBERTAR}}&\multirow{2}{*}{\xmark}&Per-free&2.79&72.9&67.8&4.05&61.5&55.9\\
     &&Per-based&5.05&54.4&50.0&6.80&38.8&34.6\\\cline{2-9} &\cmark&Per-based&\textbf{4.66}&\textbf{59.8}&\textbf{54.4}&\textbf{6.53}&\textbf{42.9}&\textbf{37.4}\\
     \specialrule{.1em}{.05em}{.05em}
		\end{tabular}}}
		\vspace{-0.2cm}
\end{table*}

\subsection{Quantitative Results on R2R}
\label{Quantitative Results-VLN}

\subsubsection{Comparison with the State-of-the-art VLN Approaches}
Table~\ref{tab:com with sota} shows the comparison of PROPER with existing VLN methods on R2R.
From Table~\ref{tab:com with sota} we can observe that  PROPER achieves consistent superiority on both validation and test sets when plugged into different state-of-the-art VLN baselines. For example, when combining with our PROPER, the improvements of the SR metric on Val Unseen for EnvDropout \cite{tan2019learning}, RelGraph \cite{hong2020language} and VLN$\circlearrowright$BERT \cite{Hong2021VLNBERTAR} are 3\%, 1\% and 1\%, respectively,
And the improvements of SPL on Test Unseen for three baselines are 4\%, 1\%, and 1\%, respectively. 
These results reveal that in perturbation-free scenes, the agent may also deviate from the correct route by a wrong action prediction, especially in unseen scenes. Our PROPER, which improves the navigation capability under deviation, can potentially improve the perturbation-free navigation robustness as well.

\subsubsection{Ablation Study}
Table~\ref{tab:effectiveness of contrastive} gives a detailed ablation study results of PROPER.
``ML'' represents using only imitation learning (IL) and reinforcement learning (RL) paradigms for training. ``CL'' means introducing the contrastive learning mechanism into training. 
For the comparing methods, 1) ``Teacher'' means only using teacher-forcing paradigm for training, 2) ``Teacher2Student'' means a four-phase curriculum learning scheme which increasingly moves from teacher-forcing to student-forcing. Concretely, in the first phase, we train the agent using pure teacher-forcing. In the second phase, we use teacher-forcing at the first four navigation timesteps, and use student-forcing at the remaining navigation timesteps. In the third phase, we use teacher-forcing at the first two navigation timesteps, and use student-forcing at the remaining navigation timesteps. In the last phase, we use pure student-forcing. 3) ``Teacher+RL'' means using teacher-forcing and RL paradigms for training, 4) ``Non-pro Aug'' represents the non-progressive variant implemented by directly requiring agents to learn to navigate under different perturbed trajectories, 5) ``Progressive Aug'' represents using the proposed progressive trajectory augmentation strategy without the perturbation-aware contrastive learning paradigm, and 6) PROPER means the whole model.

By comparing ``Progressive Aug'', ``Teacher+RL'', and ``Teacher2Student'', we can find that our proposed trajectory augmentation method is superior to these training strategy augmentation methods which are constrained with original data. By essentially introducing a large amount of new data samples, our trajectory augmentation strategy  significantly enhances the navigation ability.
Moreover, the comparison between ``Non-pro Aug'' and ``Progressive Aug'' shows that our progressive trajectory augmentation strategy can lead to more sufficient learning of perturbed trajectory and thus improve the navigation performance. 
The comparison between  ``Progressive Aug'' and ``PROPER'' further shows the effectiveness of the proposed perturbation-aware contrastive learning paradigm. 
The superiority of PROPER beyond all counterparts shows the effectiveness of each component in our method.

\begin{figure}
	\setlength{\abovecaptionskip}{3pt}
	\setlength{\belowcaptionskip}{3pt}
	\centering
	\renewcommand{\figurename}{Figure}
	\subfloat[Val seen scene]{
		\includegraphics[width=4cm,height=3cm]{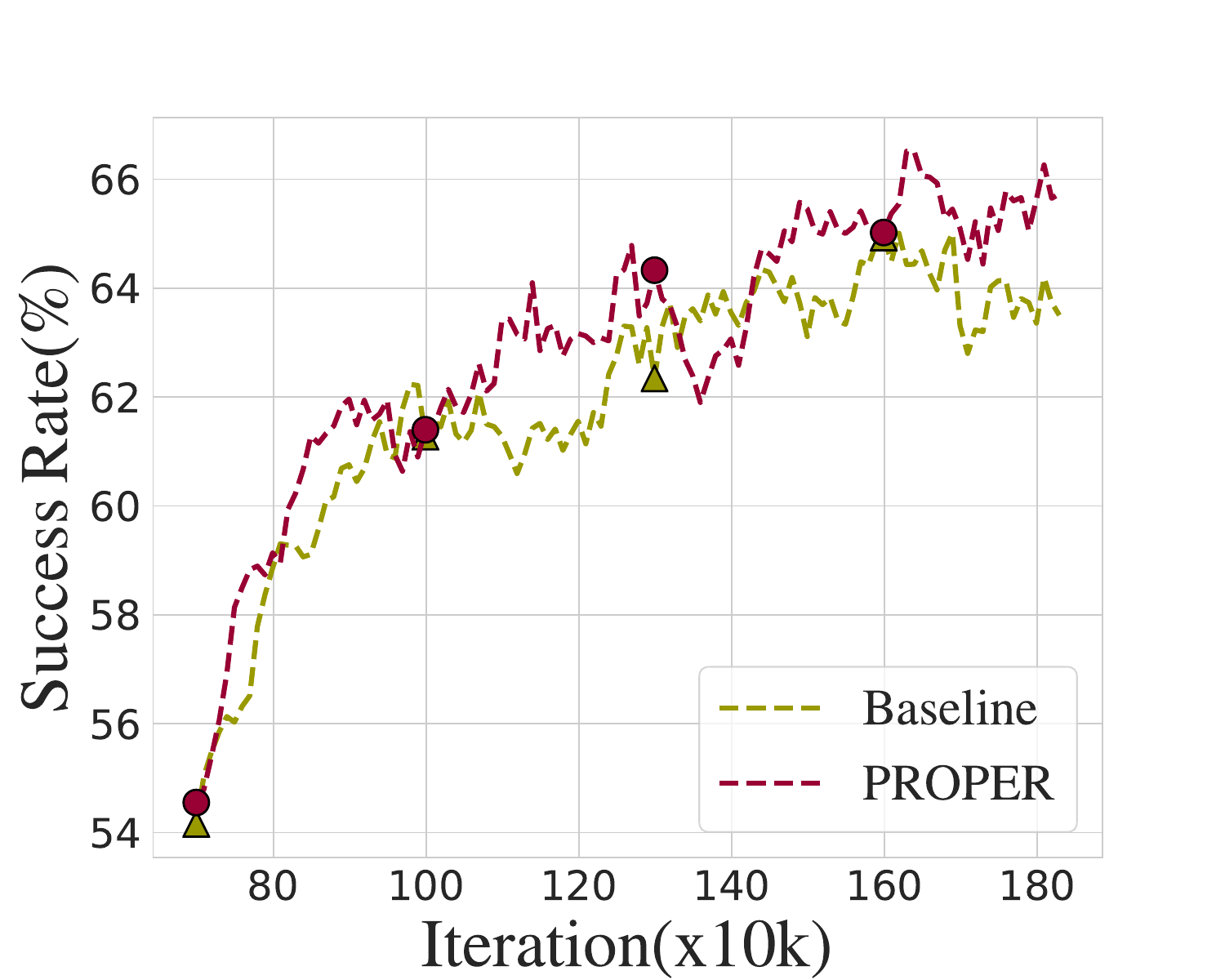}}
	\hspace{-0.01in}
	\subfloat[Val unseen scene]{
		\includegraphics[width=4cm,height=3cm]{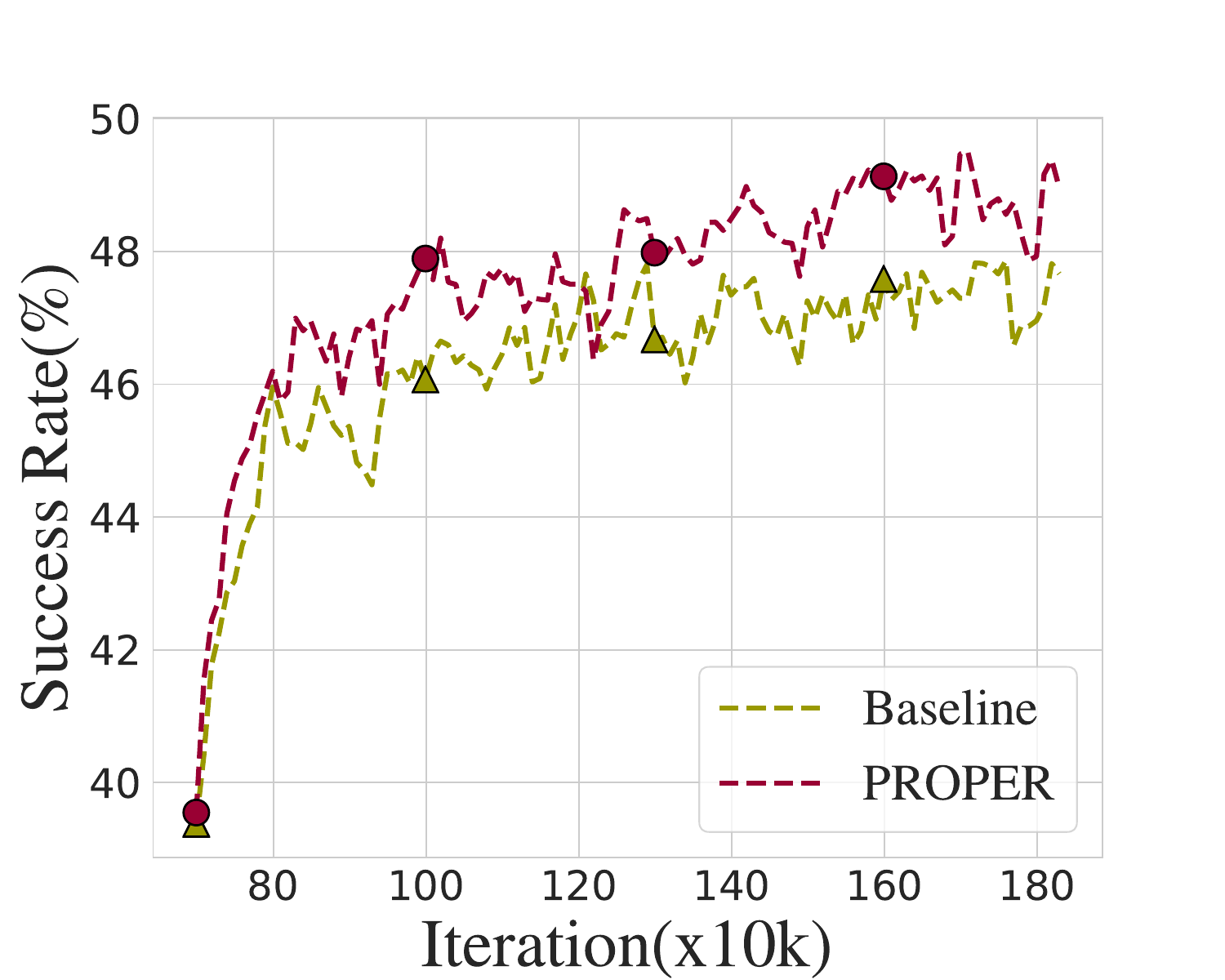}}
	
	\caption{The learning curves of PROPER and the non-progressive data augmentation baseline.}
		\vspace{-0.4cm}
	\label{fig:learning curve} 
\end{figure}

We further compare the learning curves between PROPER and the non-progressive trajectory augmentation baseline in Fig. \ref{fig:learning curve}. From Fig. \ref{fig:learning curve} we can find that 
the navigation performance can be boosted more quickly during training using PROPER than the non-progressive baseline, 
demonstrating that our proposed progressively perturbed trajectory augmentation strategy can effectively improve the training efficiency.


\begin{figure}
	\setlength{\abovecaptionskip}{3pt}
	\setlength{\belowcaptionskip}{3pt}
	\centering
	\renewcommand{\figurename}{Figure}
		
	\subfloat[Val seen scene]{
		\includegraphics[width=4cm,height=3cm]{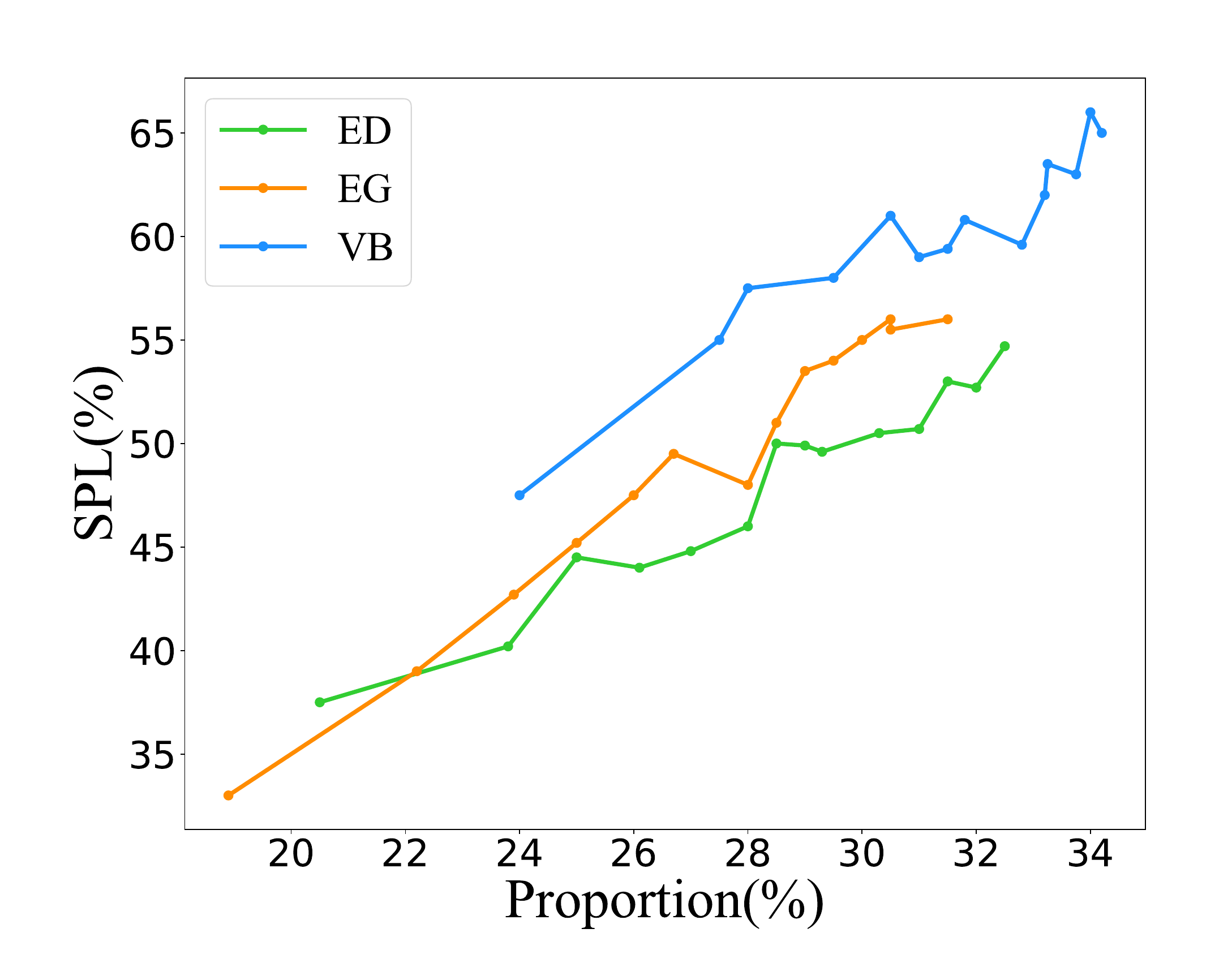}}
	\hspace{-0.01in}
	\subfloat[Val unseen scene]{
		\includegraphics[width=4cm,height=3cm]{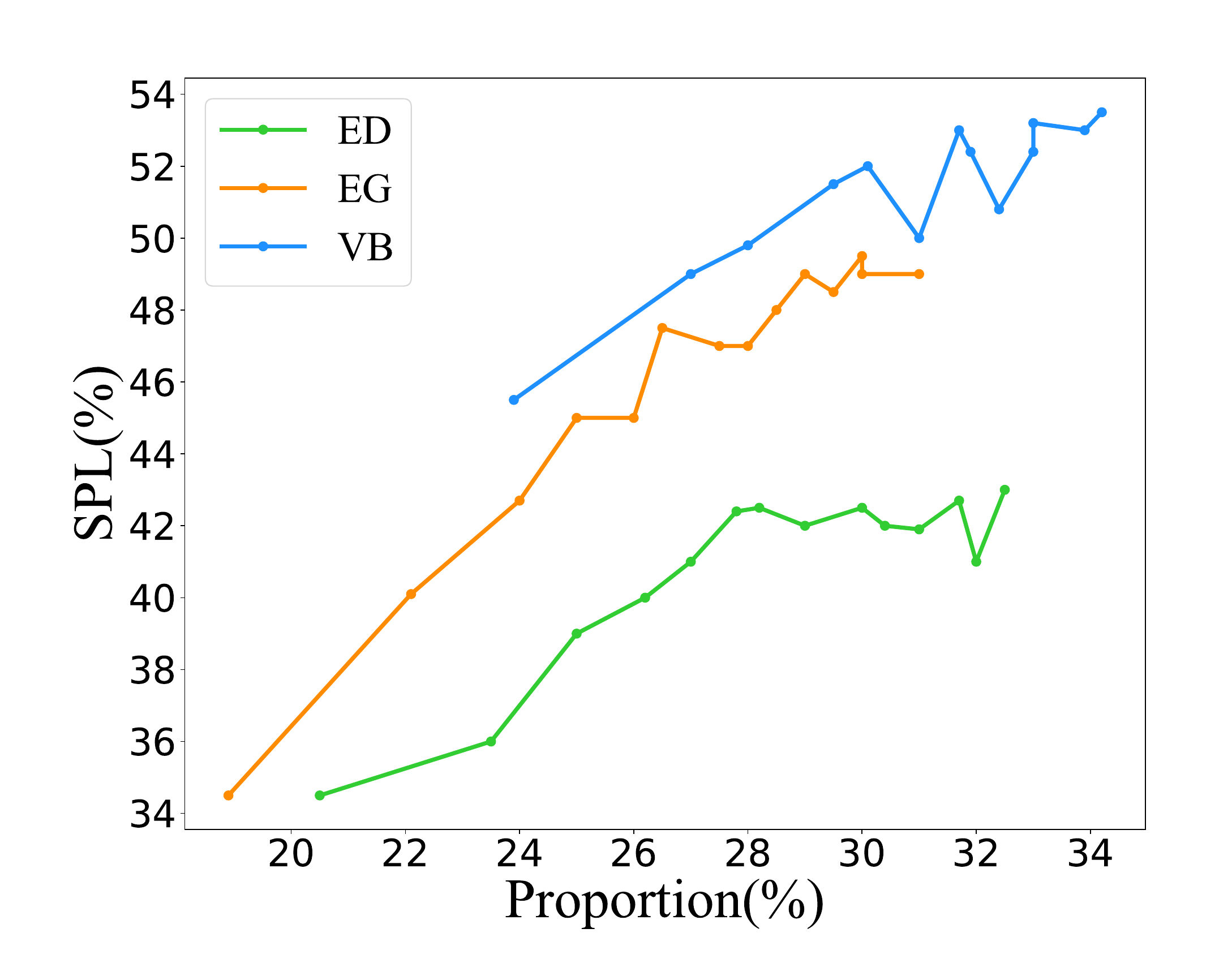}}
	
	\caption{
	SPL(\%) with respect to the perturbed data proportion on both Val Seen and Val Unseen during training.
	ED, EG, VB means EnvDropout \cite{tan2019learning}, RelGraph \cite{hong2020language} and VLN$\circlearrowright$BERT \cite{Hong2021VLNBERTAR}, respectively.}
		\vspace{-0.6cm}
	\label{fig:obs numbers} 
\end{figure}

Fig. \ref{fig:obs numbers} shows the SPL (\%) with respect to the perturbed data proportion during training. To generate Fig. \ref{fig:obs numbers}, we obtain the SPL value on the validation set and the perturbed data proportion every 2000 iterations during training. And we generate the curve based on several sampled data points. From Fig. \ref{fig:obs numbers},  we can find two important results. First, with the increasing proportion of  perturbed data, the performance (SPL) is improved. This shows that the perturbed trajectories can serve as effective augmentation data to assist navigation. Second, with the improvement of the performance (SPL), the proportion of perturbed data increases. This shows that when the agent learns to follow more ground-truth (GT) trajectories (SPL is measured based on the GT trajectory), more perturbed trajectories can be introduced. Therefore, the proposed progressive data augmentation strategy can successfully enable the agent to learn to navigate under perturbation self-adaptively with the improvement of its navigation performance for each specific trajectory.

\begin{table}[t]
	\fontsize{14}{14}\selectfont

	\caption{The evaluation results among different variants of different baseline methods in Val Unseen of PP-R2R.  ``Per-free'' and ``Per-based'' represent perturbation-free and perturbation-based environments, respectively. ED, EG, VB represent EnvDropout~\cite{tan2019learning}, RelGraph~\cite{hong2020language}, and VLN$\circlearrowright$BERT~\cite{Hong2021VLNBERTAR}, respectively.}
		
	\label{tab:effectiveness in PP-R2R}
		\resizebox{1.0\linewidth}{!}{
	{\renewcommand{\arraystretch}{1.2}	
		\begin{tabular}{c||c|c|c|c|c|c|c|c}
				\specialrule{.1em}{.05em}{.05em}	
			\hline
			\multirow{3}{*}{Method}&\multirow{3}{*}{Variant}&\multicolumn{4}{c|}{Training}&\multicolumn{3}{c}{Validation}\cr\cline{3-9}
			&&\multicolumn{2}{c|}{Per-free}&\multicolumn{2}{c|}{ Per-based}&\multicolumn{3}{c}{Per-based}\cr\cline{3-9}
			&&ML&CL&ML&CL&NE &SR &SPL\cr
			\hline
      \multirow{4}{*}{ED \cite{tan2019learning}} &1&\cmark&&&&8.44&25.9&24.3\\
 &2&&&\cmark&&\textbf{7.89}&28.3&25.5\\
          &3&\cmark&&\cmark&&8.12&28.7&24.5\\  &PROPER&\cmark&\cmark&\cmark&\cmark&7.91&\textbf{30.3}&\textbf{25.7}\\

      	\hline
		

		
		\multirow{4}{*}{EG \cite{hong2020language}} &1&\cmark&&&&8.03&28.8&27.2\\
 &2&&&\cmark&&7.66&30.6&27.1\\
          &3&\cmark&&\cmark&&7.69&30.3&25.7\\  &PROPER&\cmark&\cmark&\cmark&\cmark&\textbf{7.47}&\textbf{31.7}&\textbf{27.3}\\

      	\hline
\multirow{4}{*}{VB \cite{Hong2021VLNBERTAR}} &1&\cmark&&&&6.66&40.6&36.0\\
 &2&&&\cmark&&6.54&41.1&36.6\\
          &3&\cmark&&\cmark&&6.64&41.4&36.4\\  &PROPER&\cmark&\cmark&\cmark&\cmark&\textbf{6.53}&\textbf{42.8}&\textbf{37.4}\\

		\specialrule{.1em}{.05em}{.05em}	
		\end{tabular}}}
		\vspace{-0.4cm}
\end{table}

\begin{table*}[t]
	\fontsize{8}{8}
	\selectfont
	\caption{The navigation performance with different dialog history settings on CVDN. The baseline (PROPER with ``\xmark'') follows the architecture and the environmental dropout strategy of EnvDropout~\cite{tan2019learning}. ``Per-free'' and ``Per-based'' represent perturbation-free and perturbation-based environments, respectively. Bold fonts indicate our method.}
	\vspace{-0.2cm}
	\label{tab:results on cvdn}
		\resizebox{1.0\linewidth}{!}{
	{\renewcommand{\arraystretch}{1.2}\begin{tabular}{c||c|c|c|c|c|c|c|c}
		\specialrule{.1em}{.05em}{.05em}
			\multirow{2}{*}{Dialog history}&\multicolumn{2}{c|}{Setting}&\multicolumn{3}{c|}{Val Seen }&\multicolumn{3}{c}{Val Unseen}\cr\cline{2-9}
			&PROPER&Validation&GP(m) $\downarrow$&OSR(\%) $\uparrow$&SR(\%)  $\uparrow$&GP(m) $\downarrow$&OSR(\%) $\uparrow$&SR(\%)  $\uparrow$\cr
			\hline
            \multirow{3}{*}{Last answer}&\multirow{2}{*}{\xmark}&Per-free&5.27&72.3&64.7&3.05&52.7&36.2 \\
       
     
     &&Per-based&3.93&57.1&48.7&1.96&40.4&26.2\\ \cline{2-9}
      &\cmark&Per-based&\textbf{4.93}&\textbf{69.6}&\textbf{55.8}&\textbf{2.50}&\textbf{45.9}&\textbf{31.1}\\ 
        \hline       	
	\multirow{3}{*}{Last Q-A pair}&\multirow{2}{*}{\xmark}&Per-free&5.19&73.6&63.9&2.99&51.7&38.5\\
 &&Per-based&4.01&57.1&48.7&2.23&41.7&29.4\\ \cline{2-9}
 &\cmark&Per-based&\textbf{4.89}&\textbf{68.3}&\textbf{56.3}&\textbf{2.93}&\textbf{49.0}&\textbf{33.2}\\ \hline
     \multirow{3}{*}{All dialog history}&\multirow{2}{*}{\xmark}&Per-free&5.15&72.8&60.5&3.01&52.0&35.4\\
     &&Per-based&3.88&55.2&47.1&2.38&42.9&29.6\\\cline{2-9} &\cmark&Per-based&\textbf{4.80}&\textbf{67.0}&\textbf{54.5}&\textbf{2.97}&\textbf{45.6}&\textbf{30.8}\\
     \specialrule{.1em}{.05em}{.05em}
		\end{tabular}}}
		\vspace{-0.2cm}
\end{table*}

\subsection{Quantitative Results on PP-R2R}
\label{Quantitative Results-OVLN}

Table~\ref{tab:results on obstacle-VLN} presents the results on PP-R2R, where we can observe that
different VLN baselines show a significant performance drop compared with the results on the original dataset, demonstrating  their poor  generalization  ability to  perturbation-based environment. 
To validate the effectiveness of  PROPER in promoting perturbation-based robustness, we train these three popular VLN baselines using PROPER. 
From  Table~\ref{tab:results on obstacle-VLN} we can find that PROPER significantly promotes the perturbation-based robustness when plugged into different VLN baselines. 

To further validate the effect and necessity of each component in PROPER, we compare it with several variants which are similar to those in Table~\ref{tab:effectiveness of contrastive}. The comparison among different variants of different baseline methods is given in Table~\ref{tab:effectiveness in PP-R2R}. From Table~\ref{tab:effectiveness in PP-R2R} we can find that PROPER outperforms all counterparts in most cases, which shows that the proposed progressively perturbed trajectory augmentation strategy and perturbation-aware contrastive learning paradigm are also useful in promoting the perturbation-based navigation robustness. 




\subsection{Generalization to CVDN}
We further test the generalization ability of PROPER to another VLN-related dataset i.e., CVDN~\cite{thomason2019vision}. In CVDN, the instruction is the dialog history, which is longer and more complicated than that in R2R. The results on CVDN are presented in Table~\ref{tab:results on cvdn}. We use the standard metric GP (m) on CVDN~\cite{thomason2019vision} and two popular metrics SR (\%) and OSR (\%) on R2R~\cite{anderson2018vision} as the evaluation metrics. Following~\cite{thomason2019vision}, we test the model on different dialog history settings: Last answer, Last Q-A pair, and All dialog history.

From Table~\ref{tab:results on cvdn}, we can observe a significant performance drop consistent with that shown in Table~\ref{tab:results on obstacle-VLN} between the perturbation-free and the perturbation-based environments, which shows the effectiveness of the proposed path perturbation strategy. Moreover, we can also find that PROPER improves the perturbation-based navigation performance on CVDN largely, demonstrating the good generalization ability of PROPER.

\subsection{Qualitative Results}
\label{Qualitative Results}
\begin{figure*}
	\centering
	\includegraphics[width=0.95\linewidth]{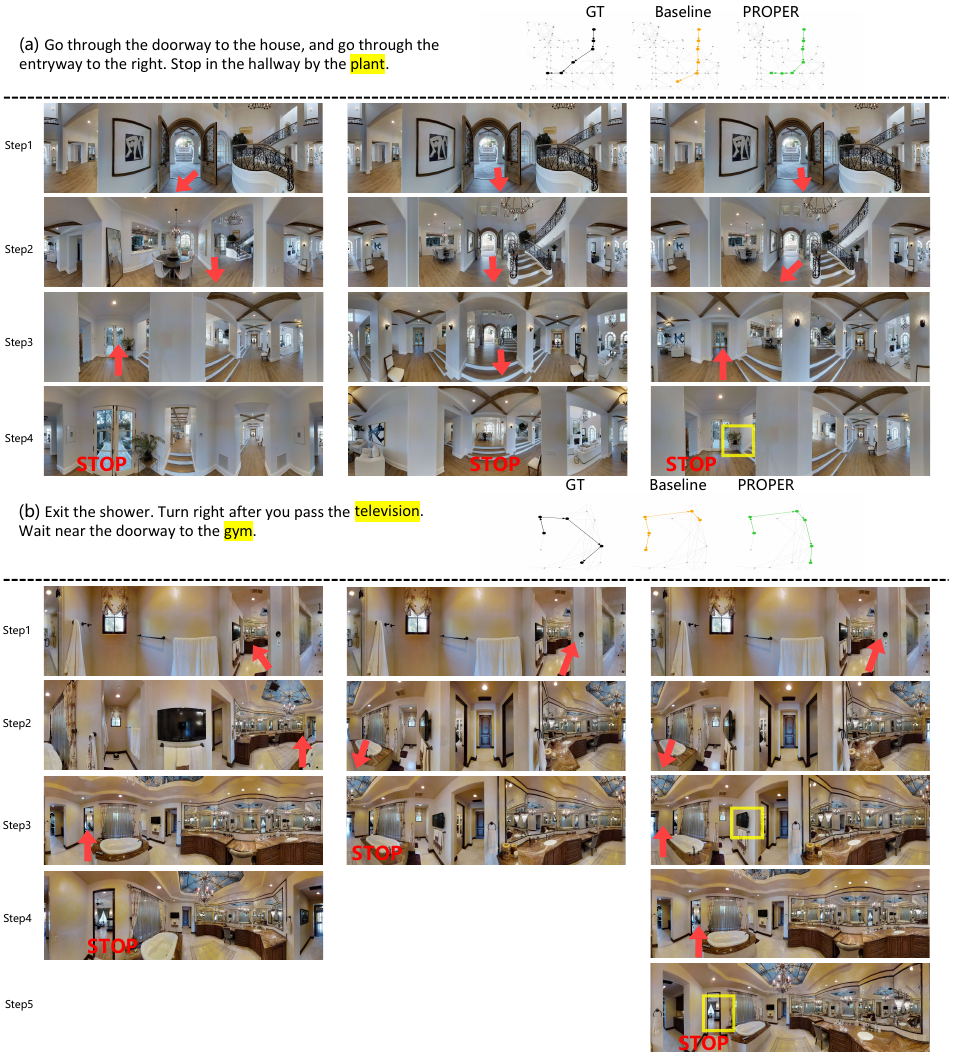}
	\caption{The visualization example under perturbation-free navigation. The agent generates the trajectory based on the instruction and panoramic view at each timestep $t$. Key visual object  is highlighted in instruction and panoramic view. In both (a) and (b), Baseline is failed and PROPER is successful according to the evaluation criteria in VLN that a trajectory is successful when the agent stops at a location  within 3 meters from the GT's ending point~\cite{anderson2018vision}.}
	
	\label{fig:visualization_per_free}
	\vspace{-0.6cm}
\end{figure*}

In this subsection, we first give some visualization results under disturbance-free navigation to validate the effectiveness of PROPER in Fig. \ref{fig:visualization_per_free}. The ground-truth (GT) trajectory, the trajectory of the baseline and the trajectory of PROPER are given for comparison. As shown in Fig. \ref{fig:visualization_per_free}, compared with the baseline, the agent trained with PROPER is capable of capturing useful multi-modal information and long-term intention of the instruction to perform successful navigation under deviation caused by the wrong action decision. For instance, in Fig. \ref{fig:visualization_per_free}(a), the agent trained with PROPER can follow the correct direction ``right'' and navigate successfully by finding the visual object ``plant'' after deviation at Step 2.
This shows that our method which improves the deviation-robustness of the agent can also benefit it in the perturbation-free environment.  

\begin{figure*}
	\centering
	\includegraphics[width=0.90\linewidth]{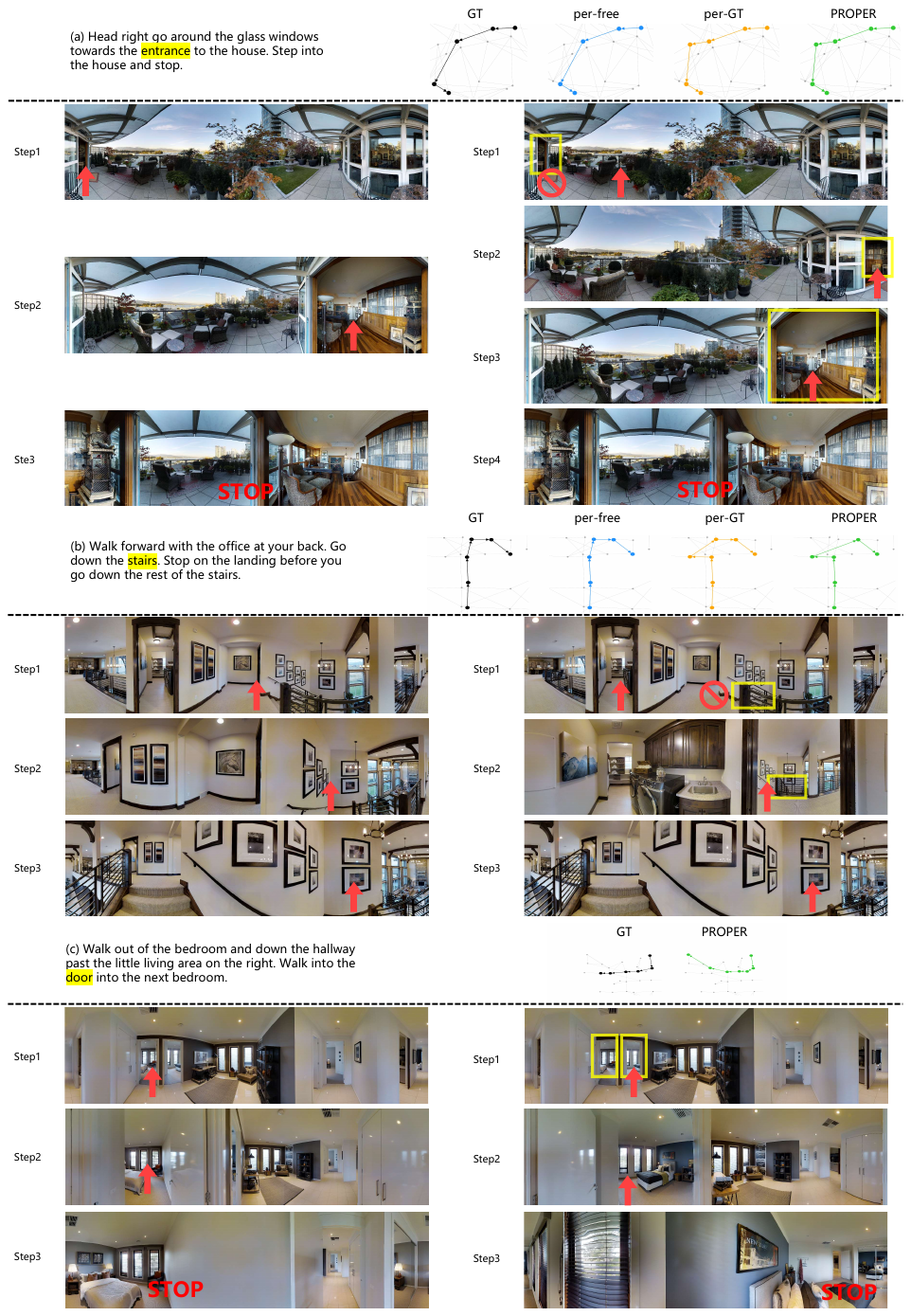}
	\vspace{-0.2cm}
	\caption{The visualization example under perturbation-based navigation. The agent generates the trajectory based on the instruction and panoramic view at each timestep $t$. Key visual object when meeting the perturbation is highlighted in both instruction and panoramic view. According to the evaluation criteria in VLN that a trajectory is successful when the agent stops at a location  within 3 meters from the GT's ending point~\cite{anderson2018vision}, PROPER is successful in both (a) and (b). And (c) is the failure case. }
	
	\label{fig:visualization}
	\vspace{-0.2cm}
\end{figure*}

Then we present qualitative results under perturbation-based navigation in Fig. \ref{fig:visualization}, to show how the perturbed path data can help the agent enhance the perturbation-based navigation robustness. The ground-truth  (GT) trajectory, perturbation-free trajectory (per-free), perturbation-aware GT trajectory (per-GT) and the trajectory of PROPER are given for comparison. We show the panoramic views at different timesteps of perturbation-free trajectory (left) and perturbation-based one under PROPER (right). From Fig. \ref{fig:visualization} we can find that by introducing perturbations, the agent is exposed to unexpected visual environment. However, due to the constraint imposed when the perturbation (see Sec.~\ref{Perturbed Trajectory Construction}), the visual environment after perturbation has a relative large overlap with that without the perturbation. And by finding the referred key visual objects in the instruction, e.g., ``entrance'' in Fig. \ref{fig:visualization}(a) and ``stairs'' in Fig. \ref{fig:visualization}(b), the agent trained with PROPER can learn to navigate back to the instruction-correlated trajectory for successful navigation. These visualization results show that PROPER effectively enhances the multi-modal understanding ability and  robustness of the navigation agent under perturbation-based scenes. We also show the failure case in Fig. \ref{fig:visualization}(c), where the agent fails since the visual observations cause ambiguity when matching the instruction (There are two doors existing in the scene). This essentially brings more difficulties in the perturbation-free environment as well, which is hard to avoid in real-world navigation scenes. We leave the solution for this in future work.


\section{Conclusion}
    \label{conclusion}
This paper proposes Progressive Perturbation-aware Contrastive Learning (PROPER) for training deviation-robust VLN agents, which introduces a simple yet effective path  perturbation scheme  into the navigation process. To better utilize the perturbed trajectory data and capture the difference brought by perturbation, a progressively perturbed trajectory augmentation strategy and a perturbation-aware contrastive learning paradigm are developed to improve the agent's robustness.  Experimental results on both the public R2R dataset and our constructed introspection subset PP-R2R show the superiority of PROPER beyond multiple state-of-the-art VLN baselines and its effectiveness in promoting navigation robustness under deviation. 

In future work, we plan to improve  the proposed method for generalizing to  more VLN benchmark datasets such as Touchdown \cite{chen2019touchdown} and REVERIE \cite{qi2020reverie}.
Promoting the navigation robustness of VLN agents under more real-world challenges, such as sensor errors or visual ambiguity when matching the instruction is also worth exploring when deploying them into real-world applications. 
 
	
	%

	

	\section*{Acknowledgments}
This work was supported in part by National Key R\&D Program of China under Grant No. 2020AAA0109700,  Guangdong Outstanding Youth Fund (Grant No. 2021B1515020061), Shenzhen Science and Technology Program (Grant No. RCYX20200714114642083) and Shenzhen Fundamental Research Program(Grant No. JCYJ20190807154211365) and Nansha Key RD Program under Grant No.2022ZD014.
 


	\ifCLASSOPTIONcaptionsoff
	\newpage
	\fi

	
	
	%
		
		
\vspace{-0.2cm}
\bibliographystyle{IEEEtran}
\bibliography{IEEEabrv,egbib}	
	
	%

\begin{IEEEbiography}[{\includegraphics[width=1in,height=1.25in,clip,keepaspectratio]{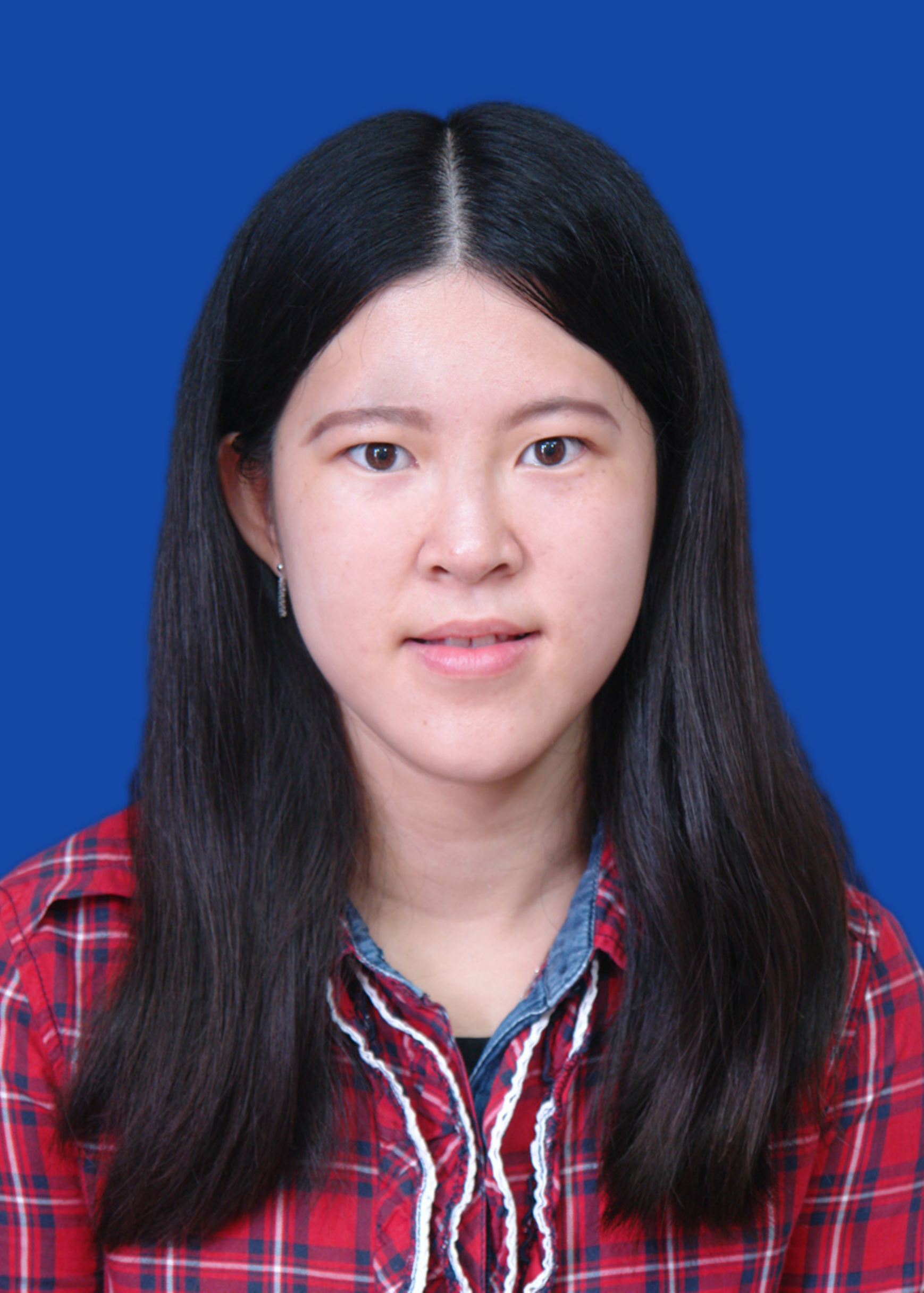}}]{Bingqian Lin} received the B.E. and the M.E.
degree in Computer Science from University of
Electronic Science and Technology of China and
Xiamen University, in 2016 and 2019, respectively.
She is currently working toward the D.Eng in the
school of intelligent systems engineering of Sun Yat-sen University. Her research interests include multimodal robot navigation and vision-language understanding.
\end{IEEEbiography}

\begin{IEEEbiography}[{\includegraphics[width=1in,height=1.25in,clip,keepaspectratio]{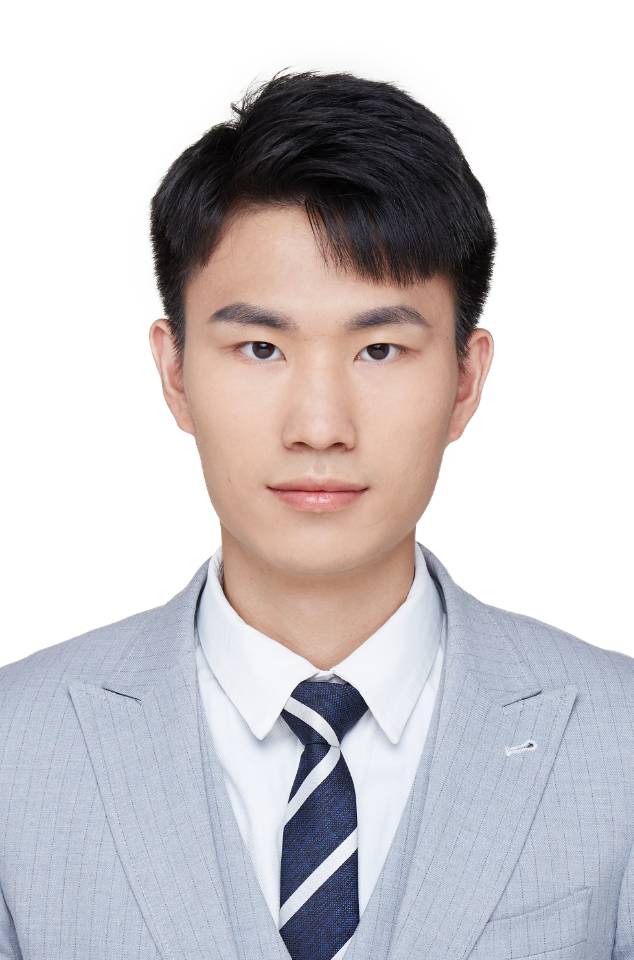}}]{Yanxin Long} is a first-year master in the School of Intelligent Systems Engineering, Sun Yat-Sen University. He works at the Human Cyber Physical Intelligence Integration Lab under the supervision of Prof. Xiaodan Liang. Before that, He received his Bachelor's Degree from the Communication College, Xidian University in 2020. His research interests  include reinforcement learning and vision-and-language understanding.
\end{IEEEbiography}

\begin{IEEEbiography}[{\includegraphics[width=1in,height=1.25in,clip,keepaspectratio]{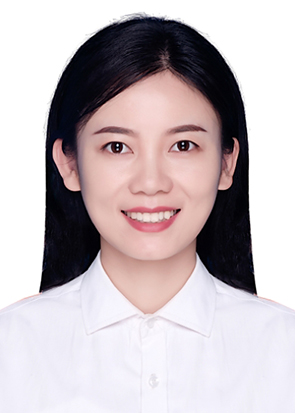}}]{Yi Zhu} received the B.S. degree in software engineering from Sun Yat-sen University, Guangzhou, China, in 2013. Since 2015, she has been a Ph.D student in computer science at the School of Electronic, Electrical, and Communication Engineering, University of Chinese Academy of Sciences, Beijing, China. Her current research interests include object recognition, scene understanding, weakly supervised learning, and visual reasoning.
\end{IEEEbiography}

\begin{IEEEbiography}[{\includegraphics[width=1in,height=1.25in,clip,keepaspectratio]{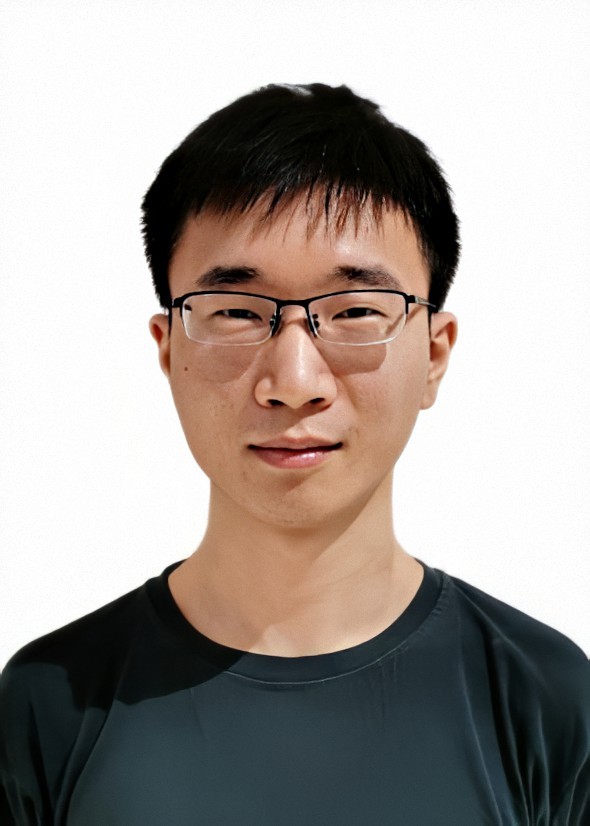}}]{Fengda Zhu} received the bachelor’s degree in School of Software Engineering from Beihang University, Beijing, China, in 2017. He is currently pursuing the Ph.D. degree with the Faculty of Information Technology, Monash University under the supervision of Prof. Xiaojun Chang. His research interests include machine learning, deep learning, and reinforcement learning.
\end{IEEEbiography}

\begin{IEEEbiography}[{\includegraphics[width=1in,height=1.25in,clip,keepaspectratio]{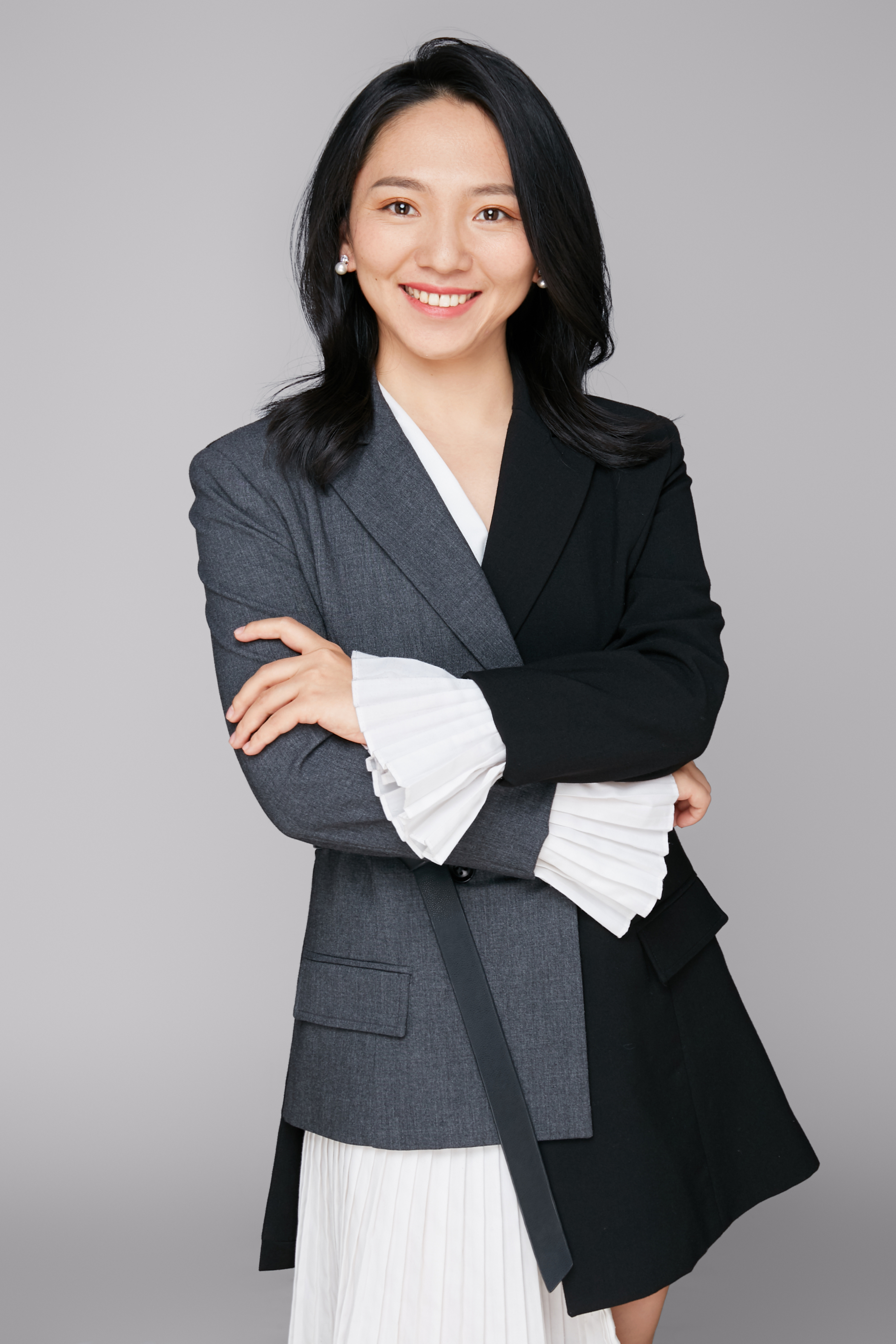}}]{Xiaodan Liang} is currently an Associate Professor at Sun Yat-sen University. She was a postdoc researcher in the machine learning department at Carnegie Mellon University, working with Prof. Eric Xing, from 2016 to 2018. She received her PhD degree from Sun Yat-sen University in 2016, advised by Liang Lin. She has published several cutting-edge projects on human-related analysis, including human parsing, pedestrian detection and instance segmentation, 2D/3D human pose estimation, and activity recognition.
\end{IEEEbiography}

\begin{IEEEbiography}[{\includegraphics[width=1in,height=1.25in,clip,keepaspectratio]{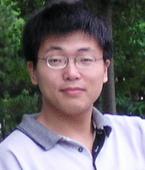}}]{Qixiang Ye} received the B.S. and M.S. degrees
in mechanical and electrical engineering from the
Harbin Institute of Technology, China, in 1999 and
2001, respectively, and the Ph.D. degree from the Institute
of Computing Technology, Chinese Academy
of Sciences, in 2006. He has been a Professor with
the University of Chinese Academy of Sciences
since 2009, and was a Visiting Assistant Professor
with the Institute of Advanced Computer Studies,
University of Maryland, College Park, in 2013. He
has authored over 50 papers in refereed conferences
and journals, and received the Sony Outstanding Paper Award. His current
research interests include image processing, visual object detection and
machine learning. He pioneered the Kernel SVM-based pyrolysis output
prediction software which was put into practical application by SINOPEC
in 2012. He developed two kinds of piecewise linear SVM methods which
were successfully applied to visual object detection.
\end{IEEEbiography}

\begin{IEEEbiography}[{\includegraphics[width=1in,height=1.25in,clip,keepaspectratio]{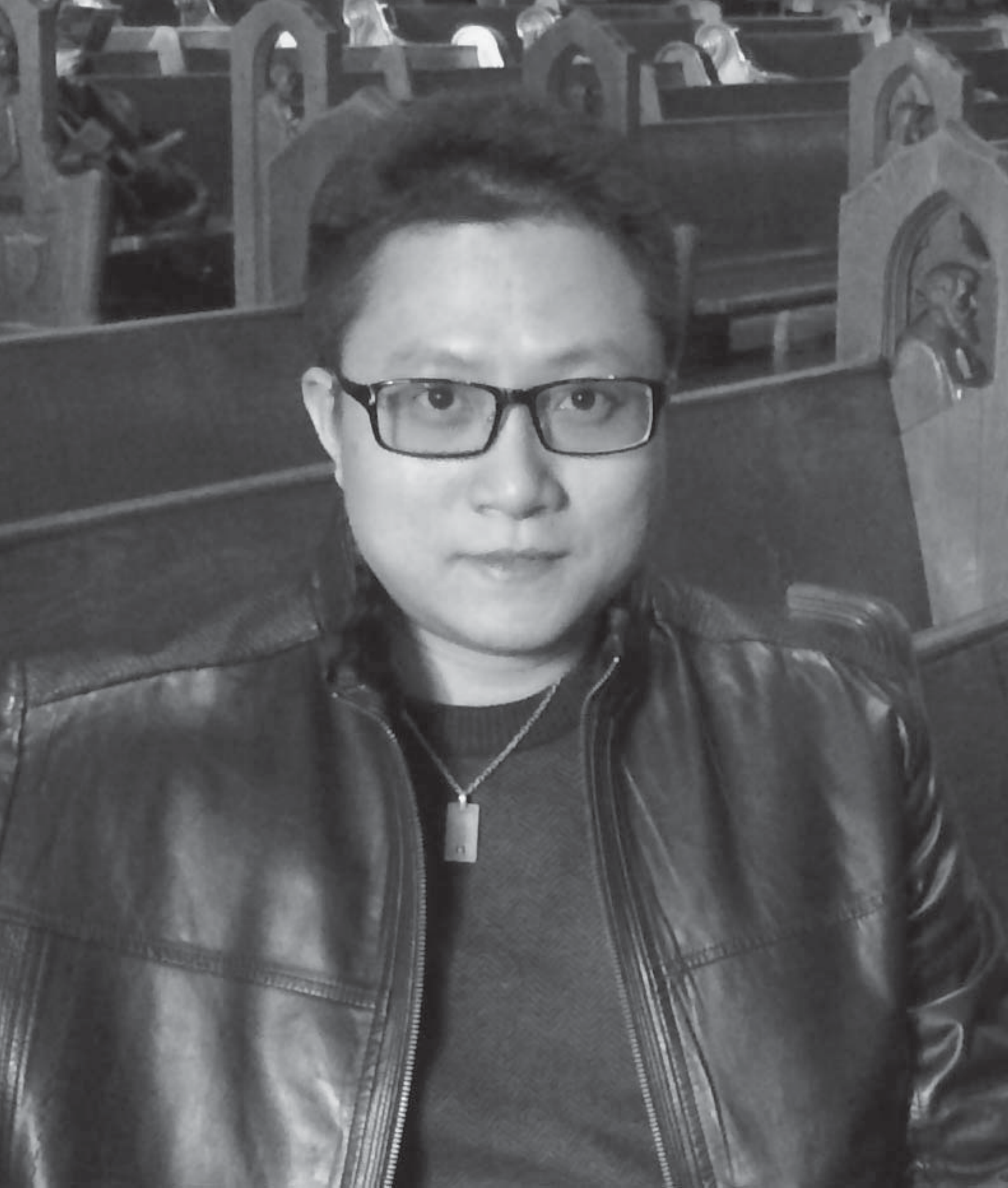}}]{Liang Lin} is CEO of DMAI Great China and a full professor of Computer Science in Sun Yat-sen University. He served as the Executive Director of the SenseTime Group from 2016 to 2018, leading the R\&D teams in developing cutting-edge, deliverable solutions in computer vision, data analysis and mining, and intelligent robotic systems.  He has authored or co-authored more than 200 papers in leading academic journals and conferences (e.g., TPAMI/IJCV, CVPR/ICCV/NIPS/ICML/AAAI). He is an associate editor of IEEE Trans, Human-Machine Systems and IET Computer Vision, and he served as the area/session chair for numerous conferences, such as CVPR, ICME, ICCV, ICMR. He was the recipient of Annual Best Paper Award by Pattern Recognition (Elsevier) in 2018, Dimond Award for best paper in IEEE ICME in 2017, ACM NPAR Best Paper Runners-Up Award in 2010, Google Faculty Award in 2012, award for the best student paper in IEEE ICME in 2014, and Hong Kong Scholars Award in 2014. He is a Fellow of IET.
\end{IEEEbiography}
	
	
	
	
	
	
	

\end{document}